\documentclass[preprint,12pt]{elsarticle}

\usepackage{amssymb}
\usepackage{amsmath}
\usepackage{amsthm}
\usepackage{url}
\usepackage[english]{babel}
\usepackage{microtype}
\usepackage{bm}
\usepackage[version=4]{mhchem}
\usepackage{booktabs}
\usepackage{multirow}
\usepackage{makecell}
\usepackage{array}
\usepackage{arydshln}
\usepackage{bbm}
\usepackage{dsfont}
\usepackage{enumitem}
\usepackage{float}
\usepackage{adjustbox}
\usepackage{subfig}
\usepackage{xcolor}
\usepackage[ruled,linesnumbered]{algorithm2e}
\graphicspath{{images/}}

\newtheorem{remark}{Remark}

\newtheorem{proposition}{Proposition}

\usepackage{lineno}

\journal{Neurocomputing}

\begin{document}

\begin{frontmatter}



% \title{Imbalance-Robust and Sampling-Efficient Continuous Conditional GANs via Adaptive Vicinity and Auxiliary Regularization}

\title{Imbalance-Robust and Sampling-Efficient Continuous Conditional GANs via Adaptive Vicinal Learning and Auxiliary Regularization}


\author[nuist,pgai]{Xin Ding\corref{cor1}}
\ead{dingxin@nuist.edu.cn}
\author[nuist,pgai]{Yun Chen}
\ead{202412491483@nuist.edu.cn}
\author[zju-media]{Yongwei Wang}
\ead{yongwei.wang@zju.edu.cn}
\author[nuist,pgai]{Kao Zhang}
\ead{kaozhang@nuist.edu.cn}
\author[nuist]{Sen Zhang}
\ead{senzhang@nuist.edu.cn}
\author[nuist,pgai]{Peibei Cao}
\ead{003970@nuist.edu.cn}
\author[nuist]{Xiangxue Wang}
\ead{xwang@nuist.edu.cn}
\author[zju-cst]{Fei Wu}
\ead{wufei@zju.edu.cn}

\cortext[cor1]{Corresponding author.}

\affiliation[nuist]{organization={School of Artificial Intelligence/School of Future Technology, Nanjing University of Information Science \& Technology},
            city={Nanjing},
            country={China}}

\affiliation[pgai]{organization={Perceptual and Generative AI Lab, Nanjing University of Information Science \& Technology},
            city={Nanjing},
            country={China}}

\affiliation[zju-media]{organization={College of Media, Zhejiang University},
            city={Hangzhou},
            country={China}}

\affiliation[zju-cst]{organization={College of Computer Science and Technology, Zhejiang University},
            city={Hangzhou},
            country={China}}

%% Abstract
\begin{abstract}
    Recent advances in continuous conditional generative modeling, including
    \textit{Continuous conditional Generative Adversarial Network} (CcGAN)
    and \textit{Continuous Conditional Diffusion Model} (CCDM), estimate
    high-dimensional data distributions conditioned on scalar regression
    labels such as angles, ages, or temperatures. However, fixed-size
    vicinal training in CcGAN can be sensitive to non-uniform label densities,
    whereas CCDM relies on computationally expensive iterative sampling. To
    address these issues, we propose CcGAN-AVAR, an imbalance-aware extension
    of CcGAN that combines soft/hybrid adaptive vicinity with auxiliary
    discriminator-guided regularization. The adaptive vicinity constructs a
    label-dependent local radius according to the available samples around
    each target condition, and the multi-task discriminator supplies both a
    regression signal for label consistency and a density-ratio-estimation
    signal for distribution matching. We further provide a theoretical
    interpretation characterizing how adaptive vicinal weighting affects the
    local bias--variance behavior of the discriminator target, how hybrid
    truncation reduces objective-level cross-condition mixing, and how the
    density-ratio-based generator penalty approximates a Pearson $\chi^2$
    discrepancy up to the estimation error of the density-ratio branch.
    Extensive experiments on four datasets, including the newly constructed
    imbalanced RC-49-I, covering resolutions from $64\times64$ to
    $256\times256$ across eleven settings, demonstrate that CcGAN-AVAR
    obtains strong generation quality and label consistency while preserving
    the one-step sampling efficiency of GANs, achieving
    300$\times$--2000$\times$ faster inference than CCDM. Our code is publicly
    available at \url{https://github.com/UBCDingXin/CcGAN-AVAR}.
\end{abstract}

%% Keywords

\end{frontmatter}

%\linenumbers

	\section{Introduction}
	
\begin{figure}[!htbp]  %[!htbp] 
		\centering
		\includegraphics[width=0.85\linewidth]{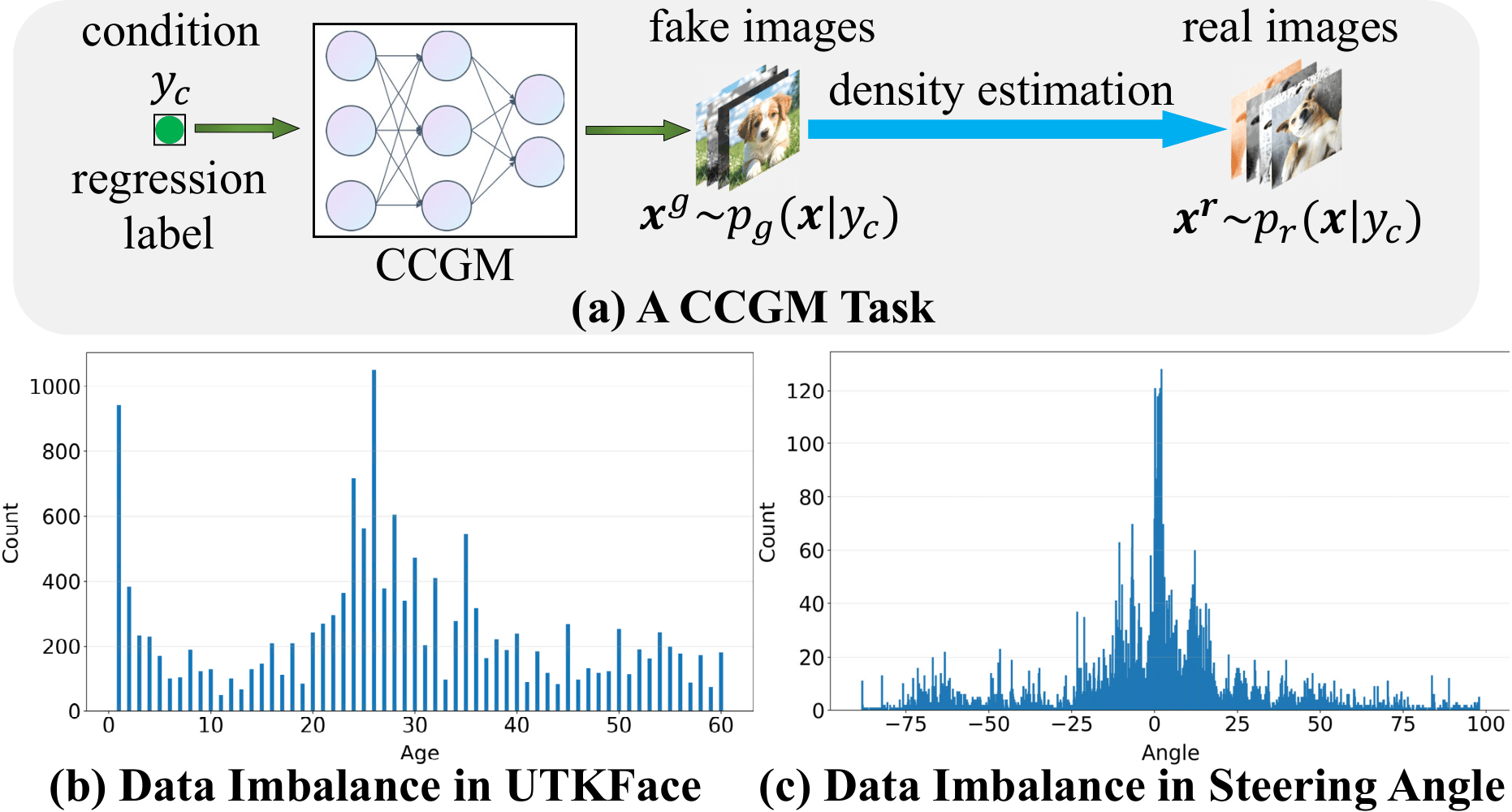}
		\caption{The illustration of CCGM with data imbalance in UTKFace and Steering Angle datasets. Fig.~\ref{fig:illustration_ccgm_imblance}(a) is adapted from Fig.~1 in \cite{ding2025ccdm}.}
		\label{fig:illustration_ccgm_imblance}
	\end{figure}

	\textit{Continuous Conditional Generative Modeling} (CCGM), illustrated in Fig.~\ref{fig:illustration_ccgm_imblance}, learns high-dimensional data distributions (typically images) conditioned on continuous scalar regression labels (e.g., angles and ages). This capability has proven valuable across diverse applications including engineering inverse design~\cite{heyrani2021pcdgan, zhao2024ccdpm, fang2023diverse}, hyperspectral image synthesis~\cite{zhu2023data}, controllable point cloud generation~\cite{triess2022point}, \ce{CO2} propagation prediction~\cite{ferreira2022predicting}, and integrated circuit design~\cite{park2022gan}. However, CCGM faces unique challenges: the inherent scarcity (or complete absence) of training samples for specific label values fundamentally limits the effectiveness of conventional \textit{conditional Generative Adversarial Networks} (cGANs)~\cite{kang2023scaling, ma2020ddcgan,xu2021conditional} and \textit{Conditional Diffusion Models} (CDMs)~\cite{peebles2023scalable, chan2024anlightendiff, zhang2025dual}.
	
	%label distribution of Steering Angle
	
	To mitigate data scarcity, Ding et al.~\cite{ding2021ccgan, ding2023ccgan, ding2022image} proposed the \textit{Continuous conditional Generative Adversarial Network} (CcGAN), which estimates the target data distribution $p_r(\bm{x}|y_c)$ by incorporating training samples within a fixed-size hard or soft vicinity of the target conditional label $y_c$. While these vicinities enable CcGAN to trade some label consistency---the degree to which generated samples align with the given condition---for improved training stability and generation quality, Fig.~\ref{fig:illustration_adaptive_vicinity}(a)--(b) reveals a critical limitation: the fixed-size vicinities may unnecessarily compromise too much label consistency for densely sampled label values, ultimately leading to suboptimal generation performance.
	
	As an alternative to CcGAN, Ding et al.~\cite{ding2025ccdm} recently integrated CcGAN techniques into diffusion models, proposing the \textit{Continuous Conditional Diffusion Model} (CCDM). Although CCDM exhibits superior generation quality and robustness to data imbalance compared to CcGAN, its sampling process is 700$\times$--2000$\times$ slower, severely limiting practical deployment. While distillation-based one-step sampling mitigates this latency, it incurs significant generation quality degradation and additional training overhead~\cite{ding2025ccdm}. Thus, achieving both high-quality generation and efficient sampling remains an open challenge in CCGM.
	
    Motivated by these limitations, we present CcGAN-AVAR, an improved CcGAN framework that targets imbalance-robust continuous conditional generation while preserving the one-step sampling advantage of GANs. Our key contributions include: 
	\begin{itemize}
		\item A Soft/Hybrid Adaptive Vicinity (SAV/HAV) mechanism that replaces fixed global vicinity parameters with a label-dependent radius, allowing dense label regions to use narrower vicinities and sparse regions to use broader vicinities.
		
		\item A multi-task discriminator with dedicated auxiliary branches for noise-insensitive regression and density ratio estimation, introducing two auxiliary regularization terms during generator training to improve label consistency and conditional distribution matching.
		
		\item A theoretical interpretation of CcGAN-AVAR. We bound the bias induced by
        vicinal weighting, relate the estimator-level conditional variance of the local
        vicinal adversarial estimate to an effective sample size, explain HAV as an
        objective-level reduction of cross-condition mixing, and connect the DRE-based
        generator penalty to an approximate Pearson $\chi^2$ discrepancy under finite
        density-ratio estimation error.
		
		\item A unified, scalable codebase for CcGAN and CcGAN-AVAR supporting multiple benchmark datasets, flexible network architectures and vicinity selection, exponential moving average, mixed-precision training, and multi-GPU acceleration.
		
		\item RC-49-I, an imbalanced variant of the RC-49 dataset~\cite{ding2023ccgan}, incorporating three distinct imbalance patterns (unimodal, bimodal, and trimodal distributions) to enable comprehensive evaluation of continuous conditional generative models under realistic data scarcity conditions. 
		
		\item The $256\times256$ versions of the UTKFace and Steering Angle datasets for validating the effectiveness of CcGAN-AVAR in more challenging high-resolution scenarios.
		
	\end{itemize}

	%%%%%%%%%%%%%%%%%%%%%%%%%%%%%%%%%%%%%%%%%%%%%%%%%%%%%%%%%%%%%%%%%%%%%%%%%
	% Preliminary
	\section{Preliminary}
	\label{sec:preliminary}
	
	%%%========================================================
	\subsection{Continuous Conditional Generative Models}
	
	To address data scarcity in CCGM tasks, CcGAN~\cite{ding2021ccgan, ding2023ccgan} leverages samples within a label-dependent vicinity of $y_c$ to approximate the data distribution $p_r(\bm{x}|y_c)$. The framework introduces two distinct vicinity formulations (hard vicinity and soft vicinity), which leads to the \textit{Hard/Soft Vicinal Discriminator Loss} (HVDL/SVDL) and the corresponding generator loss for CcGAN:
	\begin{align}
		\mathcal{L}^D_\text{adv}= & - \frac{1}{N^r}\sum_{j=1}^{N^r}\sum_{i=1}^{N^r}\mathds{E}_{\epsilon\sim\mathcal{N}(0,\sigma^2)}\left[W^r_i\log(D(\bm{x}_i^r, y_j^r+\epsilon)) \right] \nonumber\\
		&- \frac{1}{N^g}\sum_{j=1}^{N^g}\sum_{i=1}^{N^g}\mathds{E}_{\epsilon\sim\mathcal{N}(0,\sigma^2)}\left[ W^g_i\log(1-D(\bm{x}_i^g, y_j^g+\epsilon)) \right] \label{eq:vicinal_disc_loss} \\
		\mathcal{L}^G_\text{adv} =  & - \frac{1}{N^g}\sum_{i=1}^{N^g}\mathds{E}_{\epsilon\sim\mathcal{N}(0,\sigma^2)}\log (D(G(\bm{z}_{i}, y_i^g+\epsilon), y_i^g+\epsilon)), \label{eq:gen_loss}
	\end{align}
	where $D$ and $G$ denote the discriminator and generator networks, superscripts $r$/$g$ indicate real/fake samples, $W^r_i$ and $W^g_i$ are the weights for the training samples, $\bm{z}_i$ represents Gaussian noise, $\sigma$ is a positive hyperparameter, and $N^r$/$N^g$ denote real/fake sample sizes. These CcGAN losses are derived from the vanilla GAN objective~\cite{goodfellow2014generative}; their hinge loss variants are provided in Eq. (S.26) of the Appendix in \cite{ding2023ccgan}. 
	
\begin{figure}[!t]  %[!htbp] 
		\centering
		\includegraphics[width=0.85\linewidth]{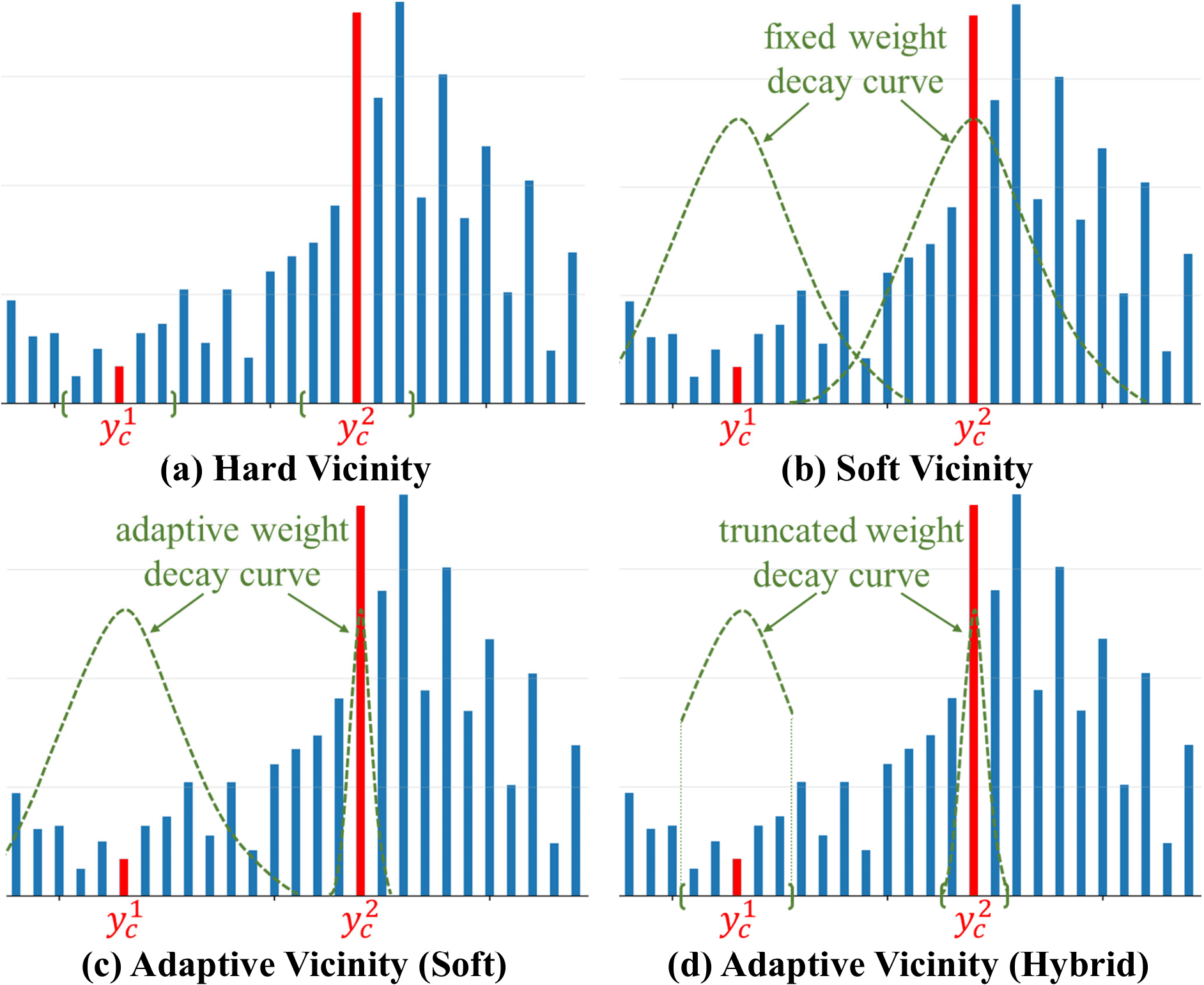}
		\caption{ \textbf{Comparison of traditional fixed-size (hard/soft) vicinities versus our proposed (soft/hybrid) adaptive vicinities.}  Bar heights represent sample sizes per label value, with $y_c^1$ and $y_c^2$ marking two conditional label values exhibiting different local sample densities.}
		\label{fig:illustration_adaptive_vicinity}
	\end{figure}

	As illustrated in Fig.~\ref{fig:illustration_adaptive_vicinity}(a), the hard vicinity defines a closed interval centered at $y_c$ with a fixed radius $\kappa$. Thus, the weights $W^r_i$ and $W^g_i$ for HVDL are defined as follows:
	\begin{equation}
		W^r_i=\frac{\mathds{1}_{\{|y_j^r+\epsilon-y_i^r|\leq\kappa\}}}{N_{y^r_j+\epsilon,\kappa}^r},\quad W^g_i=\frac{\mathds{1}_{\{|y_j^g+\epsilon-y_i^g|\leq\kappa\}}}{N_{y^g_j+\epsilon,\kappa}^g},
		\label{eq:weight_hard}
	\end{equation}
	with $\mathds{1}_{\{\cdot\}}$ as the indicator function and $N_{y_c,\kappa}$ counting samples in the hard vicinity $[y_c-\kappa,y_c+\kappa]$. Ding et al.~\cite{ding2023ccgan} provide a rule of thumb to compute $\kappa$ and $\sigma$ from samples.  For soft vicinity, all samples are used to estimate $p_r(\bm{x}|y_c)$, but each sample $\bm{x}_i$ is assigned with a weight $W_i$, with the following formulation:
	\begin{equation}
		W^r_i=\frac{w(y_i^r,y_j^r+\epsilon)}{\sum_{k=1}^{N^r}w(y_k^r,y_j^r+\epsilon)},   W^g_i=\frac{w(y_i^g,y_j^g+\epsilon)}{\sum_{k=1}^{N^g}w(y_k^g,y_j^g+\epsilon)}.
		\label{eq:weight_soft}
	\end{equation}
	
	As illustrated in Fig.~\ref{fig:illustration_adaptive_vicinity}(b), the soft vicinity weights in Eq.~\eqref{eq:weight_soft} follow an exponential decay curve
	\begin{equation}
		\label{eq:soft_exp_decay}
		w(y,y^\prime)=e^{-\nu(y-y^\prime)^2}
	\end{equation}
	where the decay rate $\nu>0$ controls the weighting sensitivity to label distances. Ding et al.~\cite{ding2023ccgan} set $\nu=1/\kappa^2$ and exclude samples with soft weights below $10^{-3}$ from discriminator updates for computational efficiency.
	
	%illustration of vicinity
	
	The conventional hard/soft vicinity (with constant $\kappa$ and $\nu$) often produces suboptimal generation quality on imbalanced datasets. As Fig.~\ref{fig:illustration_adaptive_vicinity}(a)-(b) shows, while samples cluster densely around $y_c^2$, the fixed-size vicinity formulation either incorporates excessive samples (hard vicinity) or over-weights distant samples (soft vicinity), both of which significantly degrade label consistency. 
	
	Besides the vicinal techniques, to address the challenges of integrating continuous regression labels into conditional generative models, Ding et al. introduced an \textit{Improved Label Input} (ILI) mechanism, which transforms a one-dimensional regression label $y$ into a multi-dimensional embedding vector $\bm{h}_y=\phi(y)$ via a pre-trained 5-layer multi-layer perceptron (MLP) $\phi$, enabling effective injection of the encoded label information into both the generator and discriminator networks.
	
	Recent work by \cite{ding2025ccdm} introduces CCDM, which incorporates vicinal techniques and ILI's embedding method into diffusion models as a replacement for CcGAN. Although both their empirical study and the experiments in this paper (Tables \ref{tab:main_results_low} and \ref{tab:main_results_high}) show CCDM outperforms conventional CcGANs while maintaining data imbalance robustness, its prohibitively slow sampling speed significantly limits practical applicability. To address this, \cite{ding2025ccdm} adapted a distillation-based technique \cite{yin2024improved}, proposing the one-step DMD2-M sampler. Although DMD2-M achieves significantly faster inference than CCDM's default DDIM sampler \cite{song2021denoising}, it exhibits non-negligible quality degradation, as demonstrated in \cite{ding2025ccdm}'s experiments.
	
	Thus, developing a sampling-efficient CCGM method that maintains imbalance robustness remains an open challenge.

	%%%========================================================
	\subsection{Sample-Based Density Ratio Estimation}
	
	Ding et al.~\cite{ding2023efficient} proposed cDR-RS, a sample-based \textit{Density Ratio Estimation} (DRE) method for conditional GANs. Building upon its prior work~\cite{ding2020subsampling}, the method estimates the conditional density ratio $r(\bm{x}|y)=p_r(\bm{x}|y)/p_g(\bm{x}|y)$, where $p_g(\cdot|y)$ denotes the generated/model conditional distribution induced by the generator at condition $y$. The estimated ratios are subsequently incorporated into a rejection sampling framework to improve the quality of generated samples in both class-conditional GANs and CcGANs.
	
	cDR-RS's DRE framework employs a two-stage architecture: a fixed, pre-trained encoder that maps input images $\bm{x}$ into latent representations $\bm{h}$, and a trainable 5-layer linear network $f_\text{dre}$ that processes $\bm{h}$ for density ratio estimation. During optimization, only $f_\text{dre}$ is updated using the following penalized Softplus loss~\cite{ding2020subsampling}, which is a special case of a Bregman divergence~\cite{varshney2011bayes}:
	\begin{align}
		\mathcal{L}(f_\text{dre}) =  & \frac{1}{N^g}\sum_{i=1}^{N^g}\left[ \delta(f_\text{dre}(\bm{h}_i^g|y_i^g))f_\text{dre}(\bm{h}_i^g|y_i^g)  -\eta(f_\text{dre}(\bm{h}_i^g|y_i^g)) \right] \nonumber\\
		&- \frac{1}{N^r}\sum_{i=1}^{N^r}\delta(f_\text{dre}(\bm{h}_i^r|y_i^r)) \nonumber\\
		& + \lambda_\text{dre}\cdot\left(\frac{1}{N^g}\sum_{i=1}^{N^g} f_\text{dre}(\bm{h}_i^g|y_i^g) - 1 \right)^2 \label{eq:cDR-RS_loss} 
	\end{align}
	where $\delta$ and $\eta$ denote respectively the Sigmoid and Softplus functions, and $\lambda_\text{dre}$ is a hyperparameter for regularization.

	%%%%%%%%%%%%%%%%%%%%%%%%%%%%%%%%%%%%%%%%%%%%%%%%%%%%%%%%%%%%%%%%%%%%%%%%%
	% Methodology
	\section{Method}
	\label{sec:method}
	
	%%%========================================================
	\subsection{Overview}
	\label{sec:overview}
	
\begin{figure}[!t]   %[!htbp] 
		\centering
		\includegraphics[width=1\linewidth]{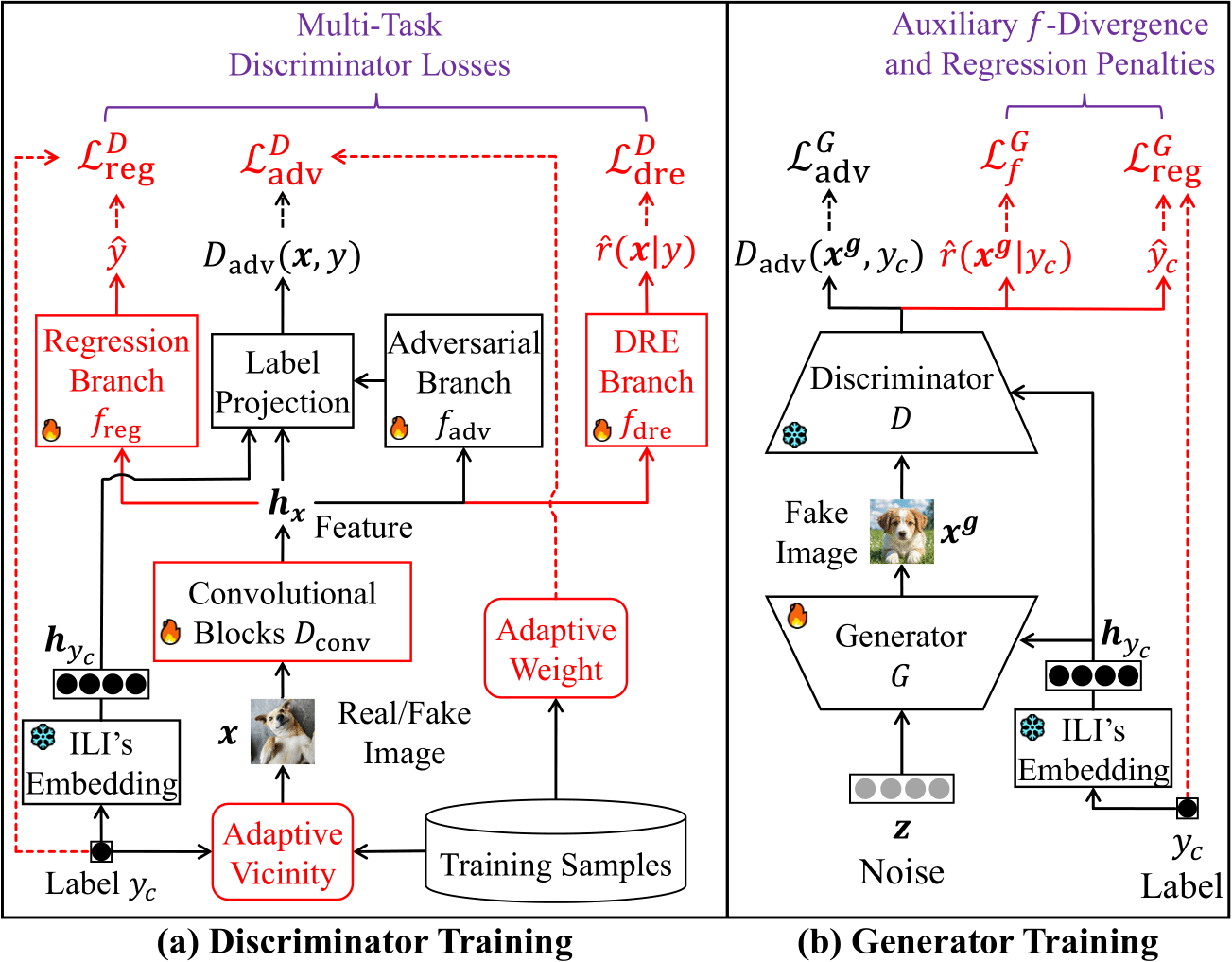}
		\caption{ \textbf{Training pipeline of the CcGAN-AVAR framework.} Novel components and procedures are highlighted in red. Data flow and loss computations are depicted by solid and dashed lines, respectively. The regression label $y_c$ is converted into an embedding vector $\bm{h}_{y_c}$ via the Improved Label Input (ILI) mechanism \cite{ding2023ccgan}. Trainable and frozen modules are distinctly marked by flame and frost icons. }
		\label{fig:training_workflow}
	\end{figure}

	In this section, we present CcGAN-AVAR, an improved CcGAN framework designed to address data imbalance challenges and serve as an efficient alternative to CCDM by enabling significantly faster inference. As illustrated in Fig.~\ref{fig:training_workflow}, our approach consists of two main algorithmic components: (1) a soft/hybrid adaptive vicinity mechanism (introduced in Section \ref{sec:AV}) that dynamically adjusts the local radius according to sample density, and (2) a multi-task discriminator (described in Section \ref{sec:aux_gen_reg}) that generates two auxiliary regularization terms to enhance generator training. Section~\ref{sec:theory_interpretation} then provides a theoretical interpretation of these components, including the bias--variance behavior of adaptive vicinal weighting, the objective-level effect of HAV, and the approximation error of the DRE-based generator penalty. The neural network architectures are detailed in Section \ref{sec:network_arch}, and a unified, scalable codebase for CcGAN-AVAR---which differs from prior CcGAN implementations \cite{ding2021ccgan, ding2023ccgan, ding2024turning, ding2022image, ding2023efficient}---is described in Section \ref{sec:codebase}.
	
	%%%% Illustrative workflows

	%%%========================================================
	\subsection{Adaptive Vicinity for Imbalanced Data}
	\label{sec:AV}
	
	Let $Q^r=\{(\bm{x}^r_i,y^r_i)\}_{i=1}^{N^r}$ denote a training set of $N^r$ real image-label pairs, where the labels consist of $M$ distinct values $Y^u=\{ y_{(1)}, ..., y_{(M)} \}$ sorted in strictly increasing order ($y_{(i)}<y_{(i+1)}$ for $i=1,...,M$). For $i$-th distinct label value $y_{(i)}$, we denote by $N_i$ its corresponding sample count. In severely imbalanced datasets, $N_i$ exhibits substantial variation across different $i$, as demonstrated in Fig.~\ref{fig:illustration_ccgm_imblance}(b)-(c).
	
	To address data imbalance, we propose two distinct types of \textit{Adaptive Vicinity} (AV): Soft AV (SAV) and Hybrid AV (HAV). Both approaches replace the fixed vicinity parameters $\kappa$ and $\nu$ in Eqs.~\eqref{eq:weight_hard} and \eqref{eq:weight_soft} with a new hyperparameter $N_\text{AV}$, which serves as the sample-count threshold for constructing a label-dependent local radius. As discussed in Section~\ref{sec:theory_interpretation}, this construction affects the resulting effective sample size and the bias--variance behavior of the vicinal discriminator target. Practical guidance for setting $N_\text{AV}$ is provided in Remark~\ref{rmk:select_Nav}.

	%%% Soft Adaptive Vicinity
	\vspace{0.2em}
	{\setlength{\parindent}{0cm} \textit{(1) Soft Adaptive Vicinity} } 
	\vspace{0.2em}
	
	Building on CcGAN's empirically demonstrated superiority of soft vicinity over hard vicinity~\cite{ding2021ccgan,ding2023ccgan}, we first propose a soft AV mechanism (Fig.~\ref{fig:illustration_adaptive_vicinity}(c)). During discriminator training, we create a target conditional label $y_c$ by randomly adding Gaussian noise to a training label (e.g., $y_j^r+\epsilon$ in Eq.~\eqref{eq:vicinal_disc_loss}). Without loss of generality, assuming $y_c$ lies between $y_{(i-1)}$ and $y_{(i)}$, we initialize a hard vicinity centered at $y_c$ with zero-radius boundaries ($\kappa_{l}=0$, $\kappa_{r}=0$) containing $N_c=0$ samples. This vicinity then extends iteratively until $N_c \geq N_\text{AV}$: at each step, the vicinity extends leftwards if the leftmost external label is closer to $y_c$ than the rightmost external label, and rightwards otherwise. Following each extension, we update either $\kappa_{l}$ or $\kappa_{r}$ to reflect the distance of the newly incorporated label to $y_c$, while incrementing $N_c$ by the sample count at this label. Upon reaching $N_c \geq N_\text{AV}$, the final radius $\kappa_{y_c} = \max(\kappa_l, \kappa_r)$ determines the local weight decay rate $\nu_{y_c} = 1/\kappa_{y_c}^2$, which yields the \textit{local exponential decay curve}:
	\begin{equation}
		w_{y_c}(y,y_c)=e^{-\nu_{y_c}(y-y_c)^2}.
		\label{eq:soft_exp_decay_y}
	\end{equation}
	Substituting Eq.~\eqref{eq:soft_exp_decay_y} into Eq.~\eqref{eq:weight_soft}, the \textit{soft adaptive weight} for the $i$-th real/fake training sample is computed as:
	\begin{equation}
		W_{i,y_c}=\frac{w_{y_c}(y_i,y_c)}{\sum_{k=1}^{N}w_{y_c}(y_k,y_c)}=\frac{e^{-\nu_{y_c}(y_i-y_c)^2}}{\sum_{k=1}^{N}e^{-\nu_{y_c}(y_k-y_c)^2}},
		\label{eq:soft_weights_y}
	\end{equation}
	where $W_{i,y_c}$ depends on both the distance of $y_i$ from $y_c$ and the specific value of $y_c$. 
	
	Replacing $W_i^r$ and $W_i^g$ in Eq.~\eqref{eq:vicinal_disc_loss} or its hinge loss variant with $W_{i,y_c}$ yields the \textit{Soft Adaptive Vicinal Discriminator Loss} (SAVDL). While SAVDL retains the form of Eq.~\eqref{eq:vicinal_disc_loss}, it fundamentally differs in constructing soft vicinities for training samples.
	
\begin{figure*}[!htbp]  %[!htbp] 
		\centering
		\includegraphics[width=0.9\linewidth]{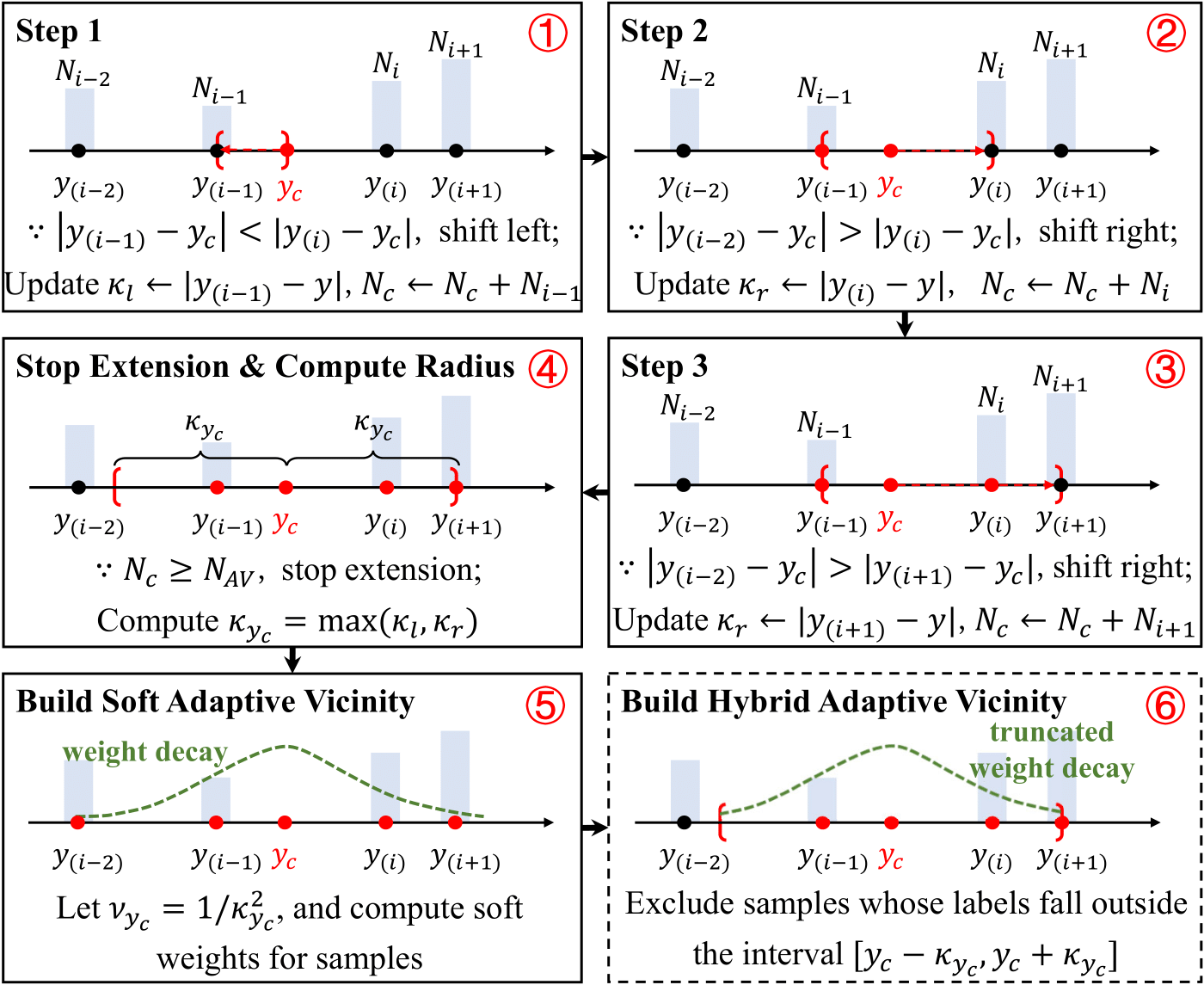}
		\caption{Illustrative example: Building the soft or hybrid adaptive vicinity for condition $y_c$ through 3 extension steps}
		\label{fig:example_building_AV}
	\end{figure*}

	As illustrated in Fig.~\ref{fig:example_building_AV}, this adaptive mechanism dynamically responds to local sample density variations: in dense regions (small $\kappa_{y_c}$), weights decay rapidly to maintain label consistency while preserving diversity; in sparse regions (large $\kappa_{y_c}$), weights decay slowly to incorporate more distant samples, ensuring training stability and preventing mode collapse.
	
	%%% Hybrid Adaptive Vicinity
	\vspace{0.2em}
	{\setlength{\parindent}{0cm} \textit{(2) Hybrid Adaptive Vicinity} } 
	\vspace{0.2em}
	
	We further propose a hybrid AV mechanism (Figs.~\ref{fig:illustration_adaptive_vicinity}(d) and \ref{fig:example_building_AV}) that extends the soft AV approach. This hybrid variant incorporates truncated weight decay by assigning zero weights to samples outside the interval $[y_c - \kappa_{y_c}, y_c + \kappa_{y_c}]$. The \textit{hybrid adaptive weight} for the $i$-th real/fake sample is computed as:
	\begin{equation}
		\label{eq:hybrid_weights_y}
		\widetilde{W}_{i,y_c}=W_{i,y_c}\cdot \mathds{1}_{\{|y_c-y_i|\leq\kappa_{y_c}\}}.
	\end{equation}
	Similarly, replacing $W_i^r$ and $W_i^g$ in Eq.~\eqref{eq:vicinal_disc_loss} or its hinge loss variant with $\widetilde{W}_{i,y_c}$ yields the \textit{Hybrid Adaptive Vicinal Discriminator Loss} (HAVDL).
	
	By combining the adaptive weighting of soft AV with the boundary constraints of hard vicinity, this approach can further enhance label consistency.
	
	\begin{remark}[Algorithm for Adaptive Vicinity Construction]
		\label{rmk:building_AV}
		Constructing the proposed soft or hybrid vicinity hinges on determining two critical parameters: the vicinity radius $\kappa_{y_c}$ and the weight decay rate $\nu_{y_c}$. A detailed procedure for computing these parameters is presented in Algorithm~\ref{alg:building_adaptive_vicinity}. 
	\end{remark}
	
	\begin{algorithm}[!htbp]
		\caption{Soft/Hybrid Adaptive Vicinity Construction for Conditional Label $y_c$. Without loss of generality, we assume $y_c$ lies between $y_{(i-1)}$ and $y_{(i)}$.}
		\label{alg:building_adaptive_vicinity} 
		\small 
		\KwData{Distinct training labels $Y^u=\{ y_{(1)}, ..., y_{(M)} \}$; Sample count for each distinct training label $\{ N_1, ..., N_M \}$; }
		\KwIn{Conditional label $y_c$; the adaptive-vicinity sample-count threshold $N_\text{AV}$;}
		\KwResult{ Vicinity width $\kappa_{y_c}$ and weight decay rate $\nu_{y_c}$ }
		$\kappa_l\leftarrow 0$, $\kappa_r\leftarrow 0$\;
		$N_c=0$\;
		$t_l=i-1$, $t_r=i$\;
		\tcp{Vicinity extension}
		\While{$N_c<N_\text{AV}$}{		
			\uIf {$|y_{(t_l)}-y_c|<|y_{(t_r)}-y_c|$} {
				\tcp{Extend leftwards}
				$\kappa_l \leftarrow |y_{(t_l)}-y_c|$\;
				$N_c \leftarrow N_c + N_{t_l}$\;
				$t_l\leftarrow t_l-1$\;
			} \uElseIf {$|y_{(t_l)}-y_c|>|y_{(t_r)}-y_c|$} {
				\tcp{Extend rightwards}
				$\kappa_r \leftarrow |y_{(t_r)}-y_c|$\;
				$N_c \leftarrow N_c + N_{t_r}$\;
				$t_r\leftarrow t_r+1$\;
			} \Else {
				\tcp{Extend in both directions}
				$\kappa_l \leftarrow |y_{(t_l)}-y_c|$\;
				$\kappa_r \leftarrow |y_{(t_r)}-y_c|$\;
				$N_c \leftarrow N_c + N_{t_l} + N_{t_r}  $\;
				$t_l\leftarrow t_l-1$\;
				$t_r\leftarrow t_r+1$\;
			}
		}
		$\kappa_{y_c}\leftarrow\max(\kappa_l, \kappa_r)$\;
		$\nu_{y_c}\leftarrow 1/\kappa_{y_c}^2$\;
	\end{algorithm}

    \begin{remark}[Heuristic Selection of $N_\text{AV}$]
		\label{rmk:select_Nav}
		The adaptive-vicinity sample-count threshold ($N_\text{AV}$) is a key hyperparameter in CcGAN-AVAR. Ablation studies on RC-49-I and UTKFace show that the model achieves better performance when $N_\text{AV}$ is calibrated within a moderate range, as shown in Fig.~\ref{fig:ab_line_graph_sfid_vs_Nav}.
		
		In practice, we determine $N_\text{AV}$ using the following heuristic. We first identify all distinct training labels $Y^u=\{ y_{(1)}, ..., y_{(M)} \}$ and their sample counts $\mathcal{N}=\{ N_1, ..., N_M \}$. Then, we compute the summed counts for each label and its immediate right neighbor, and store them as $\mathcal{S}=\{ S_1, ..., S_{M-1} \}$. The averaged sample count is then computed as $\bar{S}=\frac{1}{M-1}\sum_{i=1}^{M-1}S_i$, and $N_\text{AV}$ is set to an integer close to $\bar{S}$, with minor adjustments according to the sparsity of the label distribution. For example, on the bimodal RC-49-I dataset where $\bar{S}\approx 25.279$, we set $N_\text{AV}=30$. For UTKFace with $\bar{S}\approx 480$, we choose $N_\text{AV}=400$. For Steering Angle, where many $S_i\in\mathcal{S}$ cluster around 2 despite $\bar{S}\approx 13$, we set $N_\text{AV}=20$ to reduce the influence of sparse label regions.
	\end{remark}

	%%%========================================================
	\subsection{Auxiliary Generator Regularization}
	\label{sec:aux_gen_reg}
	
	%%% The multi-task discriminator
	\vspace{0.2em}
	{\setlength{\parindent}{0cm} \textit{(1) A Multi-Task Discriminator}}
	\vspace{0.2em}
	
	We propose a multi-task discriminator framework $D(\bm{x},y)$, with architecture illustrated in Fig.~\ref{fig:training_workflow} (a) and detailed in Section \ref{sec:network_arch}. In our implementation, the backbone of $D$ utilizes convolutional blocks from SNGAN \cite{miyato2018spectral}, denoted as $D_\text{conv}$, which maps the input image $\bm{x}$ to extracted features $\bm{h}_{\bm{x}}=D_\text{conv}(\bm{x},y)$. This backbone is designed to be modular and can be readily replaced with other established GAN architectures such as DCGAN \cite{radford2015unsupervised}, SAGAN \cite{zhang2019self}, or BigGAN \cite{brock2018large}. The convolutional backbone is followed by three parallel linear branches: a one-layer \textbf{adversarial branch} $f_\text{adv}$, a two-layer \textbf{regression branch} $f_\text{reg}$, and a three-layer \textbf{density ratio estimation branch} $f_\text{dre}$.
	
	The adversarial output, denoted by $D_\text{adv}(\bm{x},y)$, is generated by processing the extracted features $\bm{h}_{\bm{x}}$ through the adversarial branch $f_\text{adv}$ combined with label projection \cite{miyato2018spectral}, i.e.,
	\[
		D_\text{adv}(\bm{x},y) = f_\text{act}(f_\text{adv}(\bm{h}_{\bm{x}}) + \bm{h}_{\bm{x}} \cdot \bm{h}^\prime_{y}),
	\]
	where $\bm{h}^\prime_{y}=f_\text{proj}(\bm{h}_{y})$ and $f_\text{proj}$ is a single linear layer ensuring dimensional consistency between $\bm{h}_{\bm{x}}$ and $\bm{h}^\prime_{y}$. $f_\text{act}$ represents a Sigmoid activation when using the vanilla GAN loss, or an identity mapping for the hinge loss variant. The output $D_\text{adv}(\bm{x},y)$ is subsequently used to compute the adversarial loss $\mathcal{L}^D_\text{adv}$ following Eq.~\eqref{eq:vicinal_disc_loss} or its hinge loss counterpart. The sample weights required for this computation are provided by Eq.~\eqref{eq:soft_weights_y} or Eq.~\eqref{eq:hybrid_weights_y}, corresponding to the SAVDL or HAVDL formulations, respectively.
	
    The regression branch $f_\text{reg}$ predicts the label of the input image through $\hat{y}=f_\text{reg}(D_\text{conv}(\bm{x},y))$. To optimize this branch, we employ a $\gamma$-insensitive hinge loss adapted from support vector regression \cite{murphy2022probabilistic}, formulated as:
	\begin{align}
		\mathcal{L}^D_\text{reg} &= \frac{1}{N^r}\sum_{i=1}^{N^r}\mathds{E}_{\epsilon\sim\mathcal{N}(0,\sigma^2)}\left[ \max\left( |y_i^r+\epsilon - \hat{y}_i^r | - \gamma, 0 \right) \right] \nonumber \\
		& + \frac{1}{N^g}\sum_{i=1}^{N^g} \max\left( |y_i^g - \hat{y}_i^g | - \gamma, 0 \right),
		\label{eq:disc_reg_loss}
	\end{align}
	where $y_i^r$ and $y_i^g$ denote the ground truth label for $\bm{x}_i^r$ and the conditional label used to generate $\bm{x}_i^g$, respectively, while $\hat{y}_i^r$ and $\hat{y}_i^g$ represent the corresponding prediction from $f_\text{reg}$. The elasticity parameter $\gamma\geq 0$, empirically set as the maximum radius $\kappa_{y_i^r+\epsilon}$ in a training batch, accounts for Gaussian noise $\epsilon$ in real labels and helps mitigate misalignment between $\bm{x}_i^g$ and its conditional label $y_i^g$. To address the inherent label inconsistency between $y_i^g$ and the actual ground truth label of $\bm{x}_i^g$, we substitute $y_i^g$ with $\tilde{y}^g_i$---the output of a separately trained ResNet-18~\cite{he2016deep} regressor applied to $\bm{x}_i^g$.

	The DRE branch $f_\text{dre}$ predicts the conditional density ratio $r(\bm{x}|y)= p_r(\bm{x}|y) / p_g(\bm{x}|y)$ for real and fake images through
	\begin{equation}
		\label{eq:r_hat}
		\hat{r}(\bm{x}|y) = f_\text{dre}(D_\text{conv}(\bm{x},y)).
	\end{equation}
	This branch is optimized by minimizing a penalized Softplus loss, $\mathcal{L}_\text{dre}^D$, which takes a form similar to Eq.~\eqref{eq:cDR-RS_loss}:
	\begin{align}
		\mathcal{L}^D_\text{dre} &=  \frac{1}{N^g}\sum_{i=1}^{N^g}\left[ \delta(\hat{r}(\bm{x}_i^g|y_i^g))\hat{r}(\bm{x}_i^g|y_i^g)  -\eta(\hat{r}(\bm{x}_i^g|y_i^g)) \right] \nonumber\\
		& - \frac{1}{N^r}\sum_{i=1}^{N^r}\delta(\hat{r}(\bm{x}_i^r|y_i^r)) \nonumber \\
		& + \lambda_\text{dre}\cdot\left(\frac{1}{N^g}\sum_{i=1}^{N^g} \hat{r}(\bm{x}_i^g|y_i^g) - 1 \right)^2, \label{eq:disc_dre_loss}
	\end{align}
	where $\lambda_\text{dre}=10^{-2}$.  Unlike \cite{ding2020subsampling, ding2023efficient}, our feature extractor $D_\text{conv}$ is jointly optimized with $f_\text{dre}$ when minimizing $\mathcal{L}^D_\text{dre}$.
	
	The complete discriminator loss function is defined as:
	\begin{equation}
		\label{eq:loss_disc}
		\mathcal{L}^D = \mathcal{L}^D_\text{adv} + \lambda^D_\text{reg} \cdot \mathcal{L}^D_\text{reg} + \lambda^D_\text{dre} \cdot \mathcal{L}^D_\text{dre},
	\end{equation}
	where $\lambda^D_\text{reg}\geq 0$ and $\lambda^D_\text{dre}\geq 0$ are the weighting coefficients, and $\mathcal{L}^D_\text{adv}$ denotes SAVDL/HAVDL as defined in Section \ref{sec:AV}.
	
	%%% Regularization 
	\vspace{0.2em}
	{\setlength{\parindent}{0cm} \textit{(2) Auxiliary Regularization for Generator Training}}
	\vspace{0.2em}
	
	The generator's adversarial loss $\mathcal{L}^G_\text{adv}$ (Eq.~\eqref{eq:gen_loss}) is supplemented by two additional regularization terms derived from the multi-task discriminator: a \textbf{regression penalty} and a \textbf{$f$-divergence penalty}. The regression penalty utilizes mean absolute error to quantify label consistency: 
	\begin{equation}
		\label{eq:gen_reg_penalty}
		\mathcal{L}^G_\text{reg}=\frac{1}{N^g}\sum_{i=1}^{N^g} |y_i^g - \hat{y}_i^g |.
	\end{equation}
	This formulation directly minimizes the discrepancy between $\bm{x}_i^g$ and $\hat{y}_i^g$, thereby enforcing stronger label consistency during image synthesis. The $f$-divergence penalty, implemented through Pearson $\chi^2$ divergence with $f=(t-1)^2$ \cite{nowozin2016f}, provides distributional matching:
	\begin{equation}
		\mathcal{L}^G_f=\frac{1}{N^g}\sum_{i=1}^{N^g} \left(f_\text{dre}(D_\text{conv}(\bm{x}_i^g, y_i^g)) - 1 \right)^2.
		\label{eq:gen_dre_penalty}
	\end{equation}
	This term estimates the $f$-divergence between $p_r(\bm{x}|y)$ and $p_g(\bm{x}|y)$, guiding the generator towards more realistic data synthesis. The complete generator loss function combines these components through weighted summation:
	\begin{equation}
		\label{eq:loss_gene}
		\mathcal{L}^G = \mathcal{L}^G_\text{adv} + \lambda^G_\text{reg} \cdot \mathcal{L}^G_\text{reg} + \lambda^G_{f} \cdot \mathcal{L}^G_f,
	\end{equation}
	where $\lambda^G_\text{reg}\geq 0$ and $\lambda^G_f\geq 0$ are the weighting coefficients.
	
	% \begin{remark}[Extra Illustration for $f$-Divergence Penalty]
	% 	\label{rmk:f_div_penalty}
	% 	The $f$-divergence \cite{nowozin2016f} between $p_r(\bm{x}|y)$ and $p_g(\bm{x}|y)$ is defined as:
	% 	\begin{align}
	% 		\text{Div}_f(p_r\|p_g)&=\int_{\mathcal{X}}p_g(\bm{x}|y)f\left(\frac{p_r(\bm{x}|y)}{p_g(\bm{x}|y)}\right)\mathrm{d} \bm{x}\nonumber\\
	% 		&=\mathbb{E}_{\bm{x}\sim p_g(\bm{x}|y)} \left[f\left(\frac{p_r(\bm{x}|y)}{p_g(\bm{x}|y)}\right) \right],
	% 		\label{eq:f_div}
	% 	\end{align}
	% 	where $f$ is a convex function with $f(1)=0$. If $f$ is a quadratic function of the form $f(t) = (t-1)^2$, the $f$-divergence reduces to the Pearson $\chi^2$ divergence:
	% 	\begin{align}
	% 		\text{Div}_{\chi^2}(p_r\|p_g)&=\mathbb{E}_{\bm{x}\sim p_g(\bm{x}|y)} \left[ \left( \frac{p_r(\bm{x}|y)}{p_g(\bm{x}|y)} - 1 \right)^2 \right]\nonumber\\
	% 		&=\mathbb{E}_{\bm{x}\sim p_g(\bm{x}|y)} \left[ (r(\bm{x}|y) - 1)^2 \right].
	% 		\label{eq:chi_div}
	% 	\end{align}
		
	% 	Since $\hat{r}(\bm{x}|y) = f_\text{dre}(D_\text{conv}(\bm{x},y))$ in Eq.~\eqref{eq:r_hat} is trained during the discriminator updates to estimate the conditional density ratio $r(\bm{x}|y)= p_r(\bm{x}|y) / p_g(\bm{x}|y)$, the $f$-divergence penalty defined in Eq.~\eqref{eq:gen_dre_penalty} approximates Eq.~\eqref{eq:chi_div}, and minimizing $\mathcal{L}^G_f$ encourages the generator to better match the target data distribution.
	% \end{remark}

    %%%========================================================
    \subsection{Theoretical Interpretation}
	\label{sec:theory_interpretation}

	The preceding subsections define two changes to CcGAN: adaptive vicinal weighting and auxiliary generator regularization. We now clarify why these changes are useful from the perspective of the local targets optimized during training. The goal is not to prove global convergence of GAN training. Instead, we analyze the bias and variance of the vicinal discriminator target, explain how $N_\text{AV}$ indirectly affects local statistical reliability through the resulting weight concentration, and relate the DRE-based generator penalty to a learnable Pearson $\chi^2$ discrepancy.

	\subsubsection{Fixed Vicinity Bias}

	A fixed vicinity applies the same local sampling rule to all target labels $y_c$. This is problematic in two opposite regimes. In dense regions, a fixed radius can include unnecessarily distant labels and therefore bias the discriminator target away from the real conditional distribution at $y_c$. In sparse regions, shrinking the radius too aggressively leaves too few useful local samples and makes the local target high-variance. Adaptive vicinity is designed to resolve this tension: it reduces the radius when local samples are abundant and enlarges it when the local label density is low.

	For compact notation in the analysis below, let $P_y$ denote the real conditional distribution $p_r(\cdot|y)$, and let $P_W(\cdot|y_c)=\sum_i W_i(y_c)P_{y_i}(\cdot)$ be the vicinal target distribution induced by the normalized soft adaptive weights $W_i(y_c)$, where $W_i(y_c)\geq0$ and $\sum_i W_i(y_c)=1$. Here, $W_i(y_c)$ is not a learnable network parameter; it is the normalized vicinal weight assigned to the sample or label $y_i$ when the target condition is $y_c$. This normalized-mixture notation is used for SAV-style vicinal targets; HAV truncates these weights without renormalization and is analyzed separately in Section~\ref{sec:unorm_hav_cross_cond_mixing}. For a function class $\mathcal{F}$, the \textbf{integral probability metric} (IPM) is
	\begin{equation}
		d_{\mathcal{F}}(P,Q)=\sup_{f\in\mathcal{F}}\left|\mathbb{E}_{P}f(\bm{x})-\mathbb{E}_{Q}f(\bm{x})\right|.
		\label{eq:ipm}
	\end{equation}

	Because the discriminator is trained against a vicinal target rather than the exact conditional distribution at $y_c$, using samples with labels $y_i\neq y_c$ introduces an approximation bias. The following result bounds this bias by the weighted label distance induced by the vicinal weights.
	\begin{proposition}[Bias Induced By Vicinal Weighting]
	\label{prop:vicinal_bias}
	If there exists a constant $L>0$ such that $d_{\mathcal{F}}(P_y,P_{y_c})\leq L|y-y_c|$, then
	\begin{equation}
		d_{\mathcal{F}}\left(P_W(\cdot|y_c),P_{y_c}\right)
		\leq L\sum_i W_i(y_c)|y_i-y_c|.
		\label{eq:vicinal_bias_bound}
	\end{equation}
	\end{proposition}
	The proof is given in Appendix~D. Eq.~\eqref{eq:vicinal_bias_bound} shows that using distant labels increases the bias of the discriminator target. Consequently, in dense label regions, the vicinity should shrink so that the weighted label distance $\sum_i W_i(y_c)|y_i-y_c|$ remains small; otherwise a fixed wide vicinity can improve sample count but degrade label consistency.

	\subsubsection{Adaptive Vicinity and Effective Sample Size}

	The bias bound above explains why including distant labels can distort the
    local discriminator target, but it does not describe the finite-sample
    reliability of that target. When the vicinal weights are concentrated on
    only a few samples, the resulting local adversarial estimate may have
    large sampling variability. To quantify this weight concentration after
    an adaptive vicinity has been constructed, we introduce the following
    effective sample size.
    
    Let $a=\{a_i(\cdot)\}_{i=1}^{N}$ denote the unnormalized vicinal-weight family. For a fixed target label $y_c$, let $a_i(y_c)\geq0$ be the unnormalized vicinity weight assigned to the $i$-th sample, and let $A(y_c)=\sum_{i=1}^{N} a_i(y_c)>0$. The normalized diagnostic weight is $\bar W_i(y_c)=a_i(y_c)/A(y_c)$. We define
	\begin{equation}
		N_{\mathrm{eff}}(y_c;a)
		=
		\frac{1}{\sum_{i=1}^{N} \bar W_i^2(y_c)}
		=
		\frac{\left(\sum_{i=1}^{N} a_i(y_c)\right)^2}{\sum_{i=1}^{N} a_i^2(y_c)}.
		\label{eq:neff_def}
	\end{equation}
    If $m$ samples are uniformly weighted, then
    $\sum_i\bar W_i^2(y_c)=1/m$ and hence
    $N_{\mathrm{eff}}(y_c;a)=m$. Thus, $N_{\mathrm{eff}}(y_c;a)$ measures the
    number of uniformly weighted samples that would have the same
    squared-weight concentration as the current nonuniform vicinal weights.
    
    In our setting, $N_\text{AV}$ and $N_{\mathrm{eff}}$ play different
    roles. The former is the sample-count threshold used in
    Algorithm~\ref{alg:building_adaptive_vicinity} to construct the adaptive
    vicinity, whereas the latter is computed after the weights have been
    obtained. For SAV, the training weights are already normalized, so we set
    $a_i(y_c)=W_{i,y_c}$. For HAV, we set
    $a_i(y_c)=\widetilde W_{i,y_c}$, and Eq.~\eqref{eq:neff_def} renormalizes
    the retained gated weights only for this theoretical diagnostic; HAVDL
    itself still uses the unnormalized gated weights in
    Eq.~\eqref{eq:hybrid_weights_y}. Therefore, the same $N_\text{AV}$ can
    lead to different $N_{\mathrm{eff}}(y_c;a)$ values across target labels
    and across SAV/HAV, and $N_{\mathrm{eff}}$ is not an additional training
    hyperparameter.
    
	To connect $N_{\mathrm{eff}}$ to CcGAN training, fix a target label
    $y_c$ and let $\ell_i(y_c)$ denote the scalar per-sample adversarial
    term in one real or fake vicinal branch of the discriminator objective,
    with the network parameters kept fixed. Consider the normalized local
    estimate
    \begin{equation}
        \widehat L_D(y_c)=\sum_i \bar W_i(y_c)\ell_i(y_c).
        \label{eq:local_vicinal_adv_estimate}
    \end{equation}
    
    \begin{proposition}[Estimator-Level Variance of the Local Vicinal Adversarial Estimate]
    \label{prop:neff_variance}
    Conditioned on $y_c$ and the normalized weights, assume that the terms
    $\ell_i(y_c)$ are independent and have finite variance. Then
    \begin{equation}
        \mathrm{Var}\!\left[\widehat L_D(y_c)\right]
        \leq
        \frac{\max_i \mathrm{Var}[\ell_i(y_c)]}{N_{\mathrm{eff}}(y_c;a)}.
        \label{eq:neff_variance}
    \end{equation}
    \end{proposition}
    
    The proof is given in Appendix D. Eq.~\eqref{eq:neff_variance} should be
    read as an estimator-level, conditional variance analysis: it describes
    the local weighted estimate at a fixed target label and a fixed network
    state, rather than the full stochastic dynamics of GAN training. Together
    with the bias bound in Eq.~\eqref{eq:vicinal_bias_bound}, it explains
    the role of $N_\text{AV}$ as a sample-count threshold for building the
    adaptive vicinity, while $N_{\mathrm{eff}}(y_c;a)$ is a
    post-construction diagnostic that characterizes the variance behavior
    induced by the resulting normalized weights.

	\subsubsection{Unnormalized HAV and Cross-Condition Mixing}
    \label{sec:unorm_hav_cross_cond_mixing}

	After the adaptive radius has been constructed, SAV and HAV differ in how they treat condition labels outside the radius. SAV keeps a smooth tail of small weights, whereas HAV applies the gate in Eq.~\eqref{eq:hybrid_weights_y} without renormalizing the retained weights. For a fixed target condition $y_c$, a sample paired with condition $y_i$ contributes to the local discriminator objective for $y_c$. When $|y_i-y_c|$ is large, this term represents stronger cross-condition mixing in the vicinal objective. The raw weighted condition-mismatch mass below therefore measures how much distant-condition training mass remains in the local discriminator objective. It is a diagnostic of the training objective, not a surrogate for the true label error of generated images. The next result shows that, relative to SAV, the HAV gate cannot increase this objective-level condition-mismatch mass.
	\begin{proposition}[Condition-Mismatch Reduction By HAV]
	\label{prop:hav_condition_mass}
	Let $W_i(y_c)\geq0$ and $\sum_i W_i(y_c)=1$ be the SAV weights, and let $\widetilde{W}_{i,y_c}$ be the HAV weights defined in Eq.~\eqref{eq:hybrid_weights_y}. Define the corresponding raw weighted condition-mismatch masses by
	\begin{equation}
		\mathcal{C}_{\mathrm{SAV}}(y_c)=\sum_i W_i(y_c)|y_i-y_c|,\quad
		\mathcal{C}_{\mathrm{HAV}}^{\mathrm{raw}}(y_c)=\sum_i \widetilde{W}_{i,y_c}|y_i-y_c|,
		\label{eq:hav_condition_mismatch}
	\end{equation}
	and let $Z_{\mathrm{HAV}}(y_c)=\sum_i \widetilde{W}_{i,y_c}$ denote the retained HAV weight mass. Then $0\leq Z_{\mathrm{HAV}}(y_c)\leq1$ and
	\begin{equation}
		\mathcal{C}_{\mathrm{HAV}}^{\mathrm{raw}}(y_c)\leq \mathcal{C}_{\mathrm{SAV}}(y_c).
		\label{eq:hav_condition_mismatch_bound}
	\end{equation}
	\end{proposition}
	The proof is given in Appendix~D. The inequality in Eq.~\eqref{eq:hav_condition_mismatch_bound} follows from Eq.~\eqref{eq:hybrid_weights_y}: each HAV weight is either the corresponding SAV weight or zero, and no renormalization is applied. Hence, HAV removes some distant-condition contributions instead of reassigning their mass to the remaining samples. This result should be interpreted at the level of the vicinal discriminator objective. It does not by itself prove that the generated image $G(\bm{z},y_c)$ has a lower semantic label error. Rather, it explains what HAV changes during training: the local objective for $y_c$ is less mixed with samples conditioned on distant $y_i$. This objective-level reduction is consistent with the empirical Label Score improvement observed for HAV in several settings, while $Z_{\mathrm{HAV}}(y_c)<1$ also shows why HAV may reduce retained support and can be less favorable when sparse-region stability or diversity is the main concern.

	\subsubsection{DRE-based Generator Penalty}

	The vicinal analysis above explains the discriminator target. We now connect the auxiliary DRE branch to the generator objective introduced in Eq.~\eqref{eq:gen_dre_penalty}. The key point is that the DRE branch supplies a learnable estimate of a density-ratio-based divergence rather than a post-hoc rejection-sampling score. For a convex function $f$ with $f(1)=0$, the $f$-divergence~\cite{qiao2010study, nowozin2016f} between $p_r(\bm{x}|y)$ and $p_g(\bm{x}|y)$ over the image space $\mathcal X$ is
	\begin{align}
		\text{Div}_f(p_r\|p_g)&=\int_{\mathcal{X}}p_g(\bm{x}|y)f\left(\frac{p_r(\bm{x}|y)}{p_g(\bm{x}|y)}\right)\mathrm{d} \bm{x}\nonumber\\
		&=\mathbb{E}_{\bm{x}\sim p_g(\bm{x}|y)} \left[f\left(\frac{p_r(\bm{x}|y)}{p_g(\bm{x}|y)}\right) \right].
		\label{eq:f_div}
	\end{align}
	When $f(t)=(t-1)^2$, Eq.~\eqref{eq:f_div} reduces to the Pearson $\chi^2$ divergence:
	\begin{align}
		\text{Div}_{\chi^2}(p_r\|p_g)&=\mathbb{E}_{\bm{x}\sim p_g(\bm{x}|y)} \left[ \left( \frac{p_r(\bm{x}|y)}{p_g(\bm{x}|y)} - 1 \right)^2 \right]\nonumber\\
		&=\mathbb{E}_{\bm{x}\sim p_g(\bm{x}|y)} \left[ (r(\bm{x}|y) - 1)^2 \right].
		\label{eq:chi_div}
	\end{align}    
	Since $\hat{r}(\bm{x}|y)=f_\text{dre}(D_\text{conv}(\bm{x},y))$ in Eq.~\eqref{eq:r_hat} is trained during discriminator updates to estimate $r(\bm{x}|y)=p_r(\bm{x}|y)/p_g(\bm{x}|y)$, the generator penalty in Eq.~\eqref{eq:gen_dre_penalty} is a learnable approximation to Eq.~\eqref{eq:chi_div}. Minimizing $\mathcal{L}^G_f$ therefore encourages the generator to reduce a Pearson $\chi^2$-type discrepancy from the target data distribution.

	Given this interpretation, the remaining issue is approximation error. The generator does not have access to the true conditional density ratio and instead uses the learned estimate $\hat r$ inside Eq.~\eqref{eq:gen_dre_penalty}. The following result quantifies how much the resulting generator penalty can deviate from the intended Pearson $\chi^2$ divergence in Eq.~\eqref{eq:chi_div} when $\hat r$ is imperfect.
	\begin{proposition}[Approximation Error of The DRE-Based Generator Penalty]
    \label{prop:dre_approx_error}
    Fix a condition $y$ and assume $p_r(\cdot|y)\ll p_g(\cdot|y)$, where $\ll$ denotes absolute continuity. Also assume $\mathrm{Div}_{\chi^2}(p_r\|p_g)<\infty$, so the density ratio $r(\bm{x}|y)=p_r(\bm{x}|y)/p_g(\bm{x}|y)$ is well-defined $p_g$-almost everywhere. If $\|\hat r-r\|_{L_2(p_g)}\leq \varepsilon_\text{dre}$, where $\|\cdot\|_{L_2(p_g)}$ is the $L_2$ norm under $p_g(\cdot|y)$, then the generator penalty in Eq.~\eqref{eq:gen_dre_penalty} satisfies
	\begin{equation}
		\left|
		\mathbb{E}_{p_g}\left[(\hat r-1)^2\right]
		-\mathrm{Div}_{\chi^2}(p_r\|p_g)
		\right|
		\leq
		2\varepsilon_\text{dre}\sqrt{\mathrm{Div}_{\chi^2}(p_r\|p_g)}+\varepsilon_\text{dre}^2.
		\label{eq:dre_approx_error}
	\end{equation}
	\end{proposition}
	The proof is given in Appendix D. This bound clarifies when the DRE regularizer is a reasonable surrogate for distribution matching. When the DRE estimation error is small, minimizing $\mathbb{E}_{p_g}[(\hat r-1)^2]$ approximates minimizing the intended Pearson $\chi^2$ discrepancy. When the error term $\varepsilon_\text{dre}$ is non-negligible, the generator-side regularizer may deviate from the intended divergence and provide less reliable distribution-matching signals. This motivates using a moderate $\lambda^G_f$ and monitoring the stability of the DRE branch in practice.

	%%%========================================================
	\subsection{Network Architecture Design}
	\label{sec:network_arch}
	
	CcGAN-AVAR employs a SNGAN-based architecture \cite{miyato2018spectral} for both generator and discriminator networks, with a modified number of residual blocks in the generator compared to the original implementation in \cite{ding2023ccgan}. While Ding et al.~\cite{ding2023ccgan} recommend SAGAN \cite{zhang2019self} for high-resolution scenarios, we observe limited compatibility between its self-attention mechanism and exponential moving average (EMA) \cite{karras2024analyzing}---a technique that effectively enhances SNGAN performance in our experiments.
	
\begin{figure}[!htbp]
		\centering
		\includegraphics[width=0.5\linewidth]{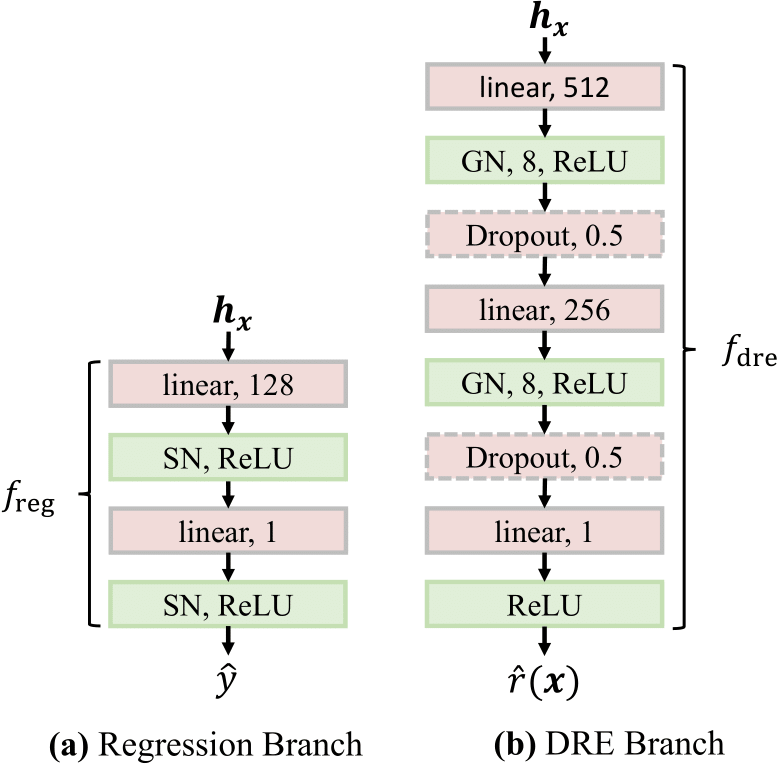} 
		\caption{\textbf{Architectures of the two-layer linear regression branch ($f_\text{reg}$) and the three-layer linear density ratio estimation (DRE) branch ($f_\text{dre}$) within the discriminator.} Spectral normalization (SN) and group normalization (GN, 8 groups) are applied in their respective components. The inclusion of Dropout in $f_\text{dre}$ is optional. The integers within the linear layers indicate the number of output channels. }
		\label{fig:disc_dre_and_reg_arch}
	\end{figure}

	A key design component is our modified discriminator architecture, which integrates both a two-layer regression branch and a three-layer density ratio estimation (DRE) branch, as shown in Fig.~\ref{fig:disc_dre_and_reg_arch}. The regression branch uses two linear layers with spectral normalization (SN) \cite{miyato2018spectral} and ReLU activation, while the DRE branch replaces SN with group normalization (GN) \cite{wu2020group} using 8 groups. In certain experimental settings, such as UTKFace ($192 \times 192$), we introduce a dropout layer after the ReLU activation. This is not for conventional overfitting prevention, but to address a specific training issue: without it, the DRE branch fails to converge, as indicated by a stagnant training loss.
	
	\begin{remark}[Relation to ACGAN]
		\label{rmk:relation_to_acgan}
		CcGAN-AVAR differs fundamentally from ACGAN \cite{odena2017conditional, kang2021rebooting, hou2022conditional} in three key aspects. First, while ACGAN uses an auxiliary classification branch for class-conditional generation, its discriminator does not directly receive class information. In contrast, CcGAN-AVAR integrates a regression branch for continuous label prediction and explicitly injects condition labels into the discriminator via label projection (Fig.~\ref{fig:training_workflow}). Second, whereas ACGAN relies on a cross-entropy loss for classification, CcGAN-AVAR employs a regression loss---specifically, a custom $\gamma$-insensitive hinge loss (Eq.~\eqref{eq:disc_reg_loss}) designed to improve robustness against label noise, rather than a standard mean squared error. Third, and most distinctively, CcGAN-AVAR implements a density ratio estimation branch that imposes an $f$-divergence penalty during generator training---a feature entirely absent in standard ACGAN and other GAN architectures.
	\end{remark}

	%%%========================================================
	\subsection{A Unified and Scalable Codebase}
	\label{sec:codebase}
	
	Prior implementations of CcGANs provided by \cite{ding2021ccgan, ding2023ccgan, ding2024turning, ding2022image, ding2023efficient} suffer from several limitations: (1) separate codebases for each benchmark dataset rather than a unified framework; (2) inconvenient adaptation to custom datasets; (3) limited support for generator and discriminator architectures (only SNGAN~\cite{miyato2018spectral} and SAGAN~\cite{zhang2019self}); (4) lack of exponential moving average (EMA)~\cite{karras2024analyzing} support; (5) absence of mixed-precision training~\cite{micikevicius2018mixed} and multi-GPU support, making them inefficient for GPUs with limited memory.
	
	To address these limitations, we introduce a unified and scalable codebase with the following features:
	
	\begin{itemize}
		
		\item A unified implementation supporting both the original CcGAN~\cite{ding2023ccgan, ding2024turning} and the proposed CcGAN-AVAR.
		
		\item A single-codebase solution for multiple benchmark datasets, designed for easy integration of new datasets via modifications to \texttt{dataset.py}.
		
		\item Flexible network architecture support, including DCGAN~\cite{radford2015unsupervised}, SNGAN~\cite{miyato2018spectral}, SAGAN~\cite{zhang2019self}, and BigGAN~\cite{brock2018large}, with configurations automatically adapted for resolutions of $64\times 64$, $128\times 128$, $192\times 192$, and $256\times 256$. This flexibility allows us to investigate the impact of the architecture on performance. Our ablation study in Section~\ref{sec:ablation} reveals that the simpler SNGAN performs best, a finding that aligns with the design rationale outlined in Section~\ref{sec:network_arch}.
		
		\item Multiple vicinity selection mechanisms: vanilla hard/soft vicinity~\cite{ding2023ccgan, ding2024turning}, and the proposed soft adaptive vicinity (SAV) and hybrid adaptive vicinity (HAV).
		
		\item Training enhancements such as EMA for improved generation quality, mixed-precision training (via Accelerate~\cite{accelerate}) for reduced memory usage, and multi-GPU training for large-scale experiments.
		
	\end{itemize}

	%%%%%%%%%%%%%%%%%%%%%%%%%%%%%%%%%%%%%%%%%%%%%%%%%%%%%%%%%%%%%%%%%%%%%%%%%
	% Experiments
	\section{Experiments}
	\label{sec:experiment}
	
	%------------------------------------------------
	\subsection{Experimental Setup}
	
	{\setlength{\parindent}{0cm}\textbf{Datasets.}} Building upon established experimental protocols \cite{ding2023ccgan, ding2024turning, ding2025ccdm}, we conduct comprehensive evaluations across multiple benchmark datasets: RC-49~\cite{ding2023ccgan} (44,051 chair images at $64\times 64$ resolution with rotation angles from $0.1^\circ$ to $89.9^\circ$ in $0.1^\circ$ increments), UTKFace~\cite{utkface} (14,723 facial images across $64\times 64$, $128\times 128$, and $192\times 192$ resolutions with age labels from 1 to 60), and Steering Angle~\cite{steeringangle} (12,271 driving images at $64\times 64$ and $128\times 128$ resolutions covering steering angles from $-80^\circ$ to $80^\circ$). For UTKFace and Steering Angle experiments, all available samples are used for both training and evaluation. The RC-49 experiments employ a curated balanced subset of 11,250 images (25 samples each for 450 distinct angle labels) for training while utilizing the complete collection of 44,051 images for comprehensive evaluation. 
	
\begin{figure}[!htbp]  %[!htbp] 
		\centering
		\includegraphics[width=1\linewidth]{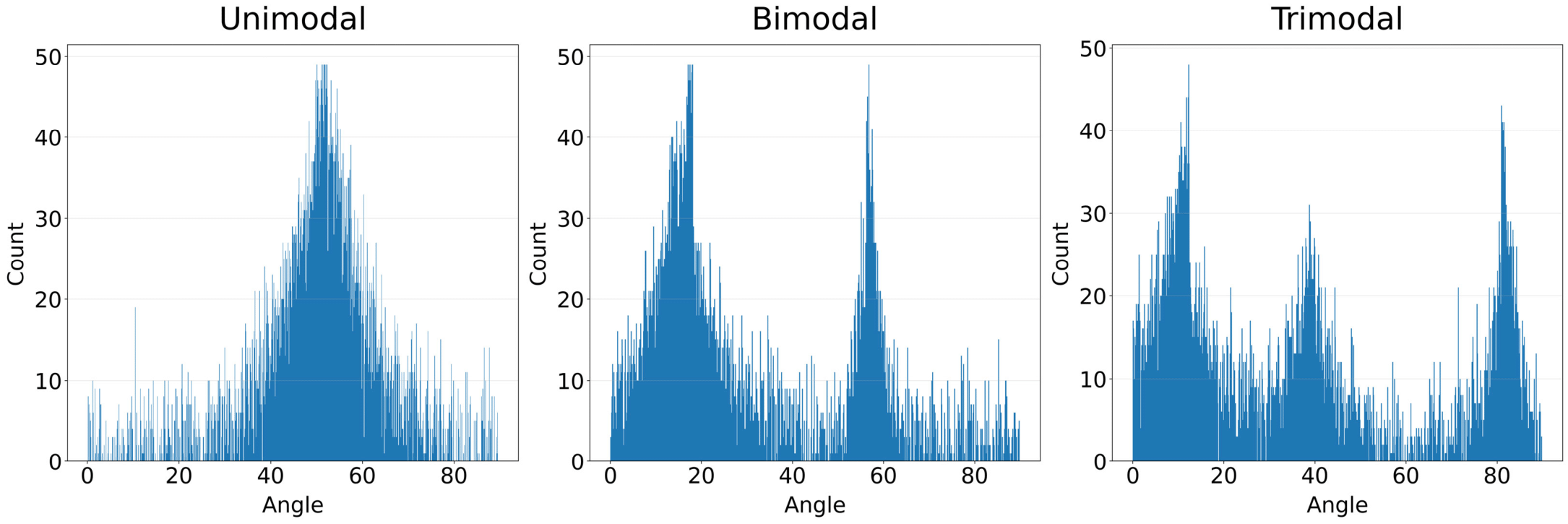}
		\caption{Three data imbalance patterns in RC-49-I}
		\label{fig:label_dist_rc64_imb}
	\end{figure}

	To rigorously evaluate model robustness to data imbalance, we introduce RC-49-I---a specialized variant of RC-49 comprising three subsets with distinct label distribution patterns (unimodal, bimodal, and trimodal; see Fig.~\ref{fig:label_dist_rc64_imb}). This framework enables systematic comparison between three challenging imbalanced scenarios (UTKFace, Steering Angle, and RC-49-I) and the balanced RC-49 baseline. All models trained on RC-49-I are evaluated using the complete RC-49 dataset (44,051 samples) to ensure consistent assessment.
	
	We also create $256\times 256$ versions of UTKFace and Steering Angle from their lower-resolution counterparts via super-resolution methods \cite{wang2021real, wang2021towards}. Detailed procedures and example images are provided in the Appendix.
	
	%label dist for RC-49-I
	
	{\setlength{\parindent}{0cm}\textbf{Compared Methods.}} Our comparative evaluation examines six continuous conditional generative models: CcDPM~\cite{zhao2024ccdpm} and CCDM~\cite{ding2025ccdm} as diffusion-based approaches; CcGAN~\cite{ding2023ccgan} and Dual-NDA~\cite{ding2024turning} representing vanilla CcGAN architectures; along with our proposed variants CcGAN-AVAR-S (soft adaptive vicinity) and CcGAN-AVAR-H (hybrid adaptive vicinity).
	
	{\setlength{\parindent}{0cm}\textbf{Implementation Setup.}} We utilize the released results of CcDPM, CCDM, CcGAN, and Dual-NDA on RC-49, UTKFace, and Steering Angle from \cite{ding2025ccdm}. For our newly introduced RC-49-I dataset, we implement these baseline methods while maintaining identical experimental configurations to those used for RC-49. For the $256\times 256$ versions of UTKFace and Steering Angle, we re-implement all methods based on their original lower-resolution setups but reduce the batch size to accommodate GPU memory constraints. 
	
	We implement CcGAN-AVAR by first establishing a strong CcGAN baseline using our unified codebase, employing SNGAN as the backbone architecture with exponential moving average and mixed-precision training enabled. This baseline achieves performance comparable to or better than the implementation in \cite{ding2023ccgan}, with results summarized in Table~\ref{tab:ablation_components} of Section~\ref{sec:ablation}. Building on this strong baseline, we develop CcGAN-AVAR-S and CcGAN-AVAR-H, setting $N_\text{AV}$ to 50 (RC-49), 30 (RC-49-I), 400 (UTKFace), and 20 (Steering Angle), and fixing hyperparameters $\lambda_\text{reg}^D=\lambda_\text{reg}^G=1$ and $\lambda_f^G=0.5$. The parameter $\lambda_\text{dre}^D$ is set to 1.0 for both variants on RC-49 and for CcGAN-AVAR-H on RC-49-I, while using 0.5 in all other cases. To compute the $\gamma$-insensitive hinge loss for the regression branch of $D$, we pre-train a regression-oriented ResNet-18 model on real image-label pairs $\{(\bm{x}^r_i,y^r_i)\}_{i=1}^{N^r}$. Complete training configurations are provided in the Appendix and our codebase.
	
	DiffAugment~\cite{zhao2020differentiable}, an effective data augmentation strategy, is applied to all GAN-based methods. Additionally, all GAN-based methods employ hinge-loss-based vicinal adversarial objectives (see Eq.~(S.26) in the Appendix of~\cite{ding2023ccgan}).
	
	{\setlength{\parindent}{0cm}\textbf{Evaluation Setup.}} For systematic evaluation, each trained model generates a predetermined number of images at specified evaluation centers across all datasets. On RC-49 and RC-49-I, models produce 200 fake images for each of 899 rotation angles spanning $0.1^\circ$ to $89.9^\circ$ in $0.1^\circ$ increments. For UTKFace, models generate 1,000 images per age from 1 to 60 years. The Steering Angle evaluation employs 2,000 uniformly distributed points across $[-80^\circ, 80^\circ]$, with 50 images synthesized per point. This protocol yields total generation counts of 179,800 images (RC-49 and RC-49-I), 60,000 images (UTKFace), and 100,000 images (Steering Angle) for each evaluated method.
	
	Following the evaluation protocol established in \cite{ding2023ccgan, ding2024turning, ding2025ccdm}, we assess generated images using one primary metric and three complementary metrics. The \textit{Sliding Fr\'echet Inception Distance} (SFID)~\cite{ding2023ccgan} serves as our primary evaluation metric, computed as the averaged FID score across all evaluation centers. Additionally, we employ three complementary metrics: (1) the \textit{Naturalness Image Quality Evaluator} (NIQE)~\cite{mittal2012making} for assessing visual realism, (2) Diversity~\cite{ding2023ccgan} for quantifying sample variety, and (3) Label Score~\cite{ding2023ccgan} for measuring label consistency. For SFID, NIQE, and Label Score, lower values indicate superior performance, whereas higher Diversity scores reflect greater sample variety. We refer interested readers to \cite{ding2025ccdm} for complete technical details regarding these evaluation metrics. 
	
	Furthermore, we evaluate the sampling efficiency of each model under a fixed sampling batch size per setting to ensure comparability. The reported metrics---sampling speed (images/second) and GPU memory consumption (GiB)---were measured on a Linux system with NVIDIA RTX 4090D.
	
	%------------------------------------------------
	\subsection{Experimental Results}
	
\begin{figure}[!htbp]
		\centering
		\includegraphics[width=0.9\linewidth]{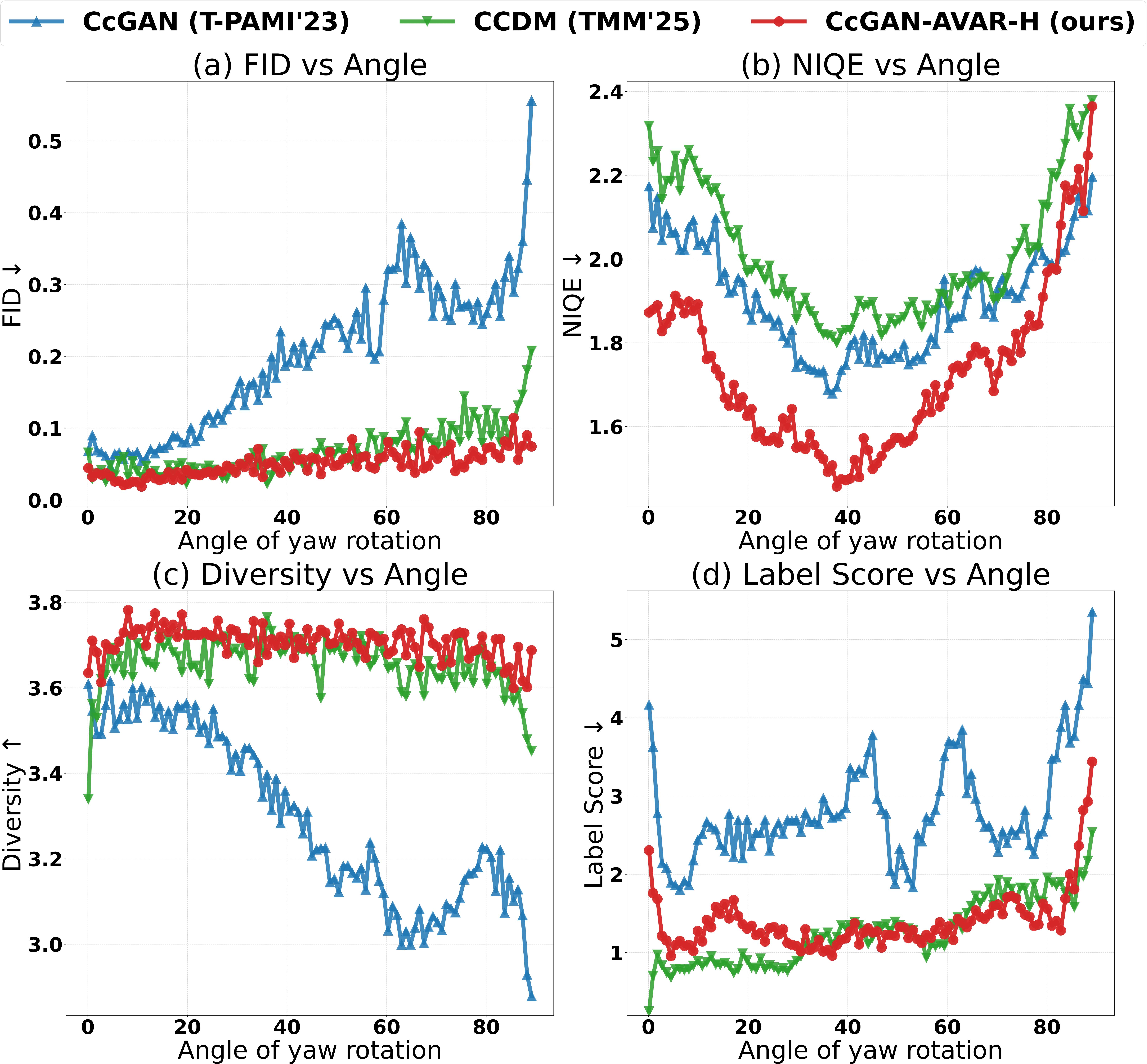} 
		\caption{ Comparison of FID/NIQE/Diversity/Label Score versus angle (evaluation center) for three compared methods on RC-49-I (bimodal). } 
		\label{fig:line_graphs_rc64_imb_bimodal}
	\end{figure}

    We conduct a comprehensive evaluation across eleven distinct settings: \textbf{ten imbalanced configurations} (RC-49-I, UTKFace, and Steering Angle from $64\times64$ to $256\times256$) and \textbf{one balanced setting} (RC-49). The empirical results, presented in Tables \ref{tab:main_results_low} and \ref{tab:main_results_high} together with Fig.~\ref{fig:line_graphs_rc64_imb_bimodal}, demonstrate that \textbf{the proposed CcGAN-AVAR framework achieves a strong balance between generation quality and sampling efficiency.} Compared to GAN-based approaches, it improves the primary SFID metric in most settings while maintaining comparable sampling efficiency. When evaluated against diffusion-based baselines, CcGAN-AVAR achieves a 300$\times$--2000$\times$ acceleration in generation time while obtaining competitive or superior generation quality in many settings. Detailed findings are shown as follows:
	
	\begin{itemize}
		\item The comparative analysis between RC-49 and RC-49-I reveals significant performance degradation for both vanilla CcGAN and its Dual-NDA variant when applied to the imbalanced RC-49-I dataset, demonstrating the substantial impact of data imbalance on CcGANs. 
		
		\item CCDM is more robust than CcGANs with a small drop of generation performance on imbalanced datasets. However, the iterative sampling process makes CCDM 700$\times$--2000$\times$ slower at sampling than CcGAN/Dual-NDA.
		
		\item The two CcGAN-AVAR variants improve the primary SFID metric in most settings, maintain comparable sampling efficiency to CcGAN/Dual-NDA, and often improve Label Score compared with GAN-based baselines, supporting the efficacy of their adaptive vicinity and auxiliary regularization designs. Crucially, CcGAN-AVAR preserves more stable generation quality between balanced (RC-49) and imbalanced (RC-49-I) datasets than vanilla CcGAN variants, indicating robustness to data skews. Furthermore, the hybrid AV variant typically outperforms soft AV in label consistency across most configurations.
		
		\item CcGAN-AVAR achieves $300\times$--$2000\times$ faster generation than CCDM while maintaining sampling speeds comparable to CcGAN and Dual-NDA, along with moderate GPU memory requirements. 
		
		\item The performance gap between CcGAN-AVAR and CCDM---in terms of both generation quality and sampling efficiency---widens at higher resolutions, most notably at $256\times 256$. In the Steering Angle ($256\times 256$) experiment, CCDM obtains a much higher Label Score than the GAN-based methods, indicating poor label control in this setting, while the CcGAN-AVAR variants achieve strong SFID and Label Score values alongside high sampling speed and low memory consumption.
		
	\end{itemize}
	
	\noindent\textbf{Metric trade-offs on UTKFace and Steering Angle.}
	Although CcGAN-AVAR performs strongly on the primary SFID metric, it does not uniformly dominate every auxiliary metric on UTKFace and Steering Angle. This behavior is expected because these datasets contain additional sources of variation beyond the scalar condition. In UTKFace, chronological age is only a noisy proxy for visual age, and images with the same age can differ substantially in identity, pose, illumination, and facial attributes. Stronger label-consistency regularization can therefore improve SFID or Label Score while slightly reducing diversity or worsening NIQE in some resolutions. In Steering Angle, the visual scene is affected by road geometry, lighting, camera viewpoint, and surrounding objects; neighboring steering angles are not always associated with smoothly varying image distributions. As a result, the label-space smoothness assumption behind vicinal training is weaker, and improvements in label consistency may come with trade-offs in NIQE or Diversity. We therefore regard SFID as the primary metric and interpret NIQE, Diversity, and Label Score as complementary diagnostics rather than requiring one method to dominate all of them simultaneously.
	
	\noindent\textbf{SAV versus HAV.}
	Tables~\ref{tab:main_results_low} and \ref{tab:main_results_high} also show that HAV is often slightly better than SAV in SFID or Label Score, but the trend is not universal. The reason is consistent with the analysis in Section~\ref{sec:theory_interpretation}: HAV removes samples outside the adaptive radius and thus reduces objective-level cross-condition mixing, which can improve label consistency when the conditional distribution changes smoothly with the label. SAV, in contrast, keeps small nonzero weights for farther labels, which can provide more support in extremely sparse or noisy label regions and may help stability or diversity. Therefore, HAV is generally preferred when label neighborhoods are moderately reliable, while SAV can be a safer choice when the label distribution is very sparse or the label-image relationship is noisy.
	
	\begin{table}[!htbp]%[htbp] 
		\setlength{\tabcolsep}{1mm}
		\footnotesize
		\centering
			
			\caption{ \textbf{Comparison of generation quality and sampling efficiency at $64\times 64$ resolution across six compared methods.} Generation quality is evaluated by SFID (overall quality), NIQE (visual realism), Diversity (sample diversity), and Label Score (label consistency). Sampling efficiency is measured by inference speed and GPU memory consumption on a single NVIDIA RTX 4090D-24G GPU. ``$\downarrow$" (``$\uparrow$") indicates lower (higher) values are preferred. The best and second-best results are highlighted in bold and underlined, respectively.}
			
			\begin{adjustbox}{max totalsize={\textwidth}{1\textheight},center}
			\begin{tabular}{cc cccc cc}
				\toprule
				\begin{tabular}[c]{@{}c@{}} \textbf{Dataset} \\ (resolution)\end{tabular} & \textbf{Method} & \begin{tabular}[c]{@{}c@{}} \textbf{SFID} \\ {(primary)} \end{tabular} $\downarrow$   & \textbf{NIQE} $\downarrow$ & \textbf{Diversity} $\uparrow$ & \begin{tabular}[c]{@{}c@{}} \textbf{Label} \\ \textbf{Score} \end{tabular} $\downarrow$ & \begin{tabular}[c]{@{}c@{}} \textbf{Speed} \\ {(img/sec)}\end{tabular} $\uparrow$ & \begin{tabular}[c]{@{}c@{}} \textbf{Memory} \\ {(GiB)} \end{tabular} $\downarrow$ \\
				\midrule
				
				%% RC-49
				\multirow{6}[1]{*}{\begin{tabular}[c]{@{}c@{}} \textbf{RC-49} \\ {(64$\times$64)}\end{tabular}} & CcDPM (AAAI'24) & 0.970 & 2.153 & 3.581 & 24.174 & 2.92 & 5.07 \\
				& CCDM (TMM'26) & 0.049 & 2.086 & 3.698 & \textbf{1.074} & 2.92 & 5.53 \\
				\cdashline{2-8}
				\specialrule{0em}{1pt}{1pt}
				& CcGAN (T-PAMI'23) & {0.126} & {1.809}  & {3.451} & 2.655 & \textbf{6243.06} & \textbf{2.62} \\
				& Dual-NDA (AAAI'24) & 0.148 & 1.808 & 3.344 & 2.211 & \textbf{6243.06} & \textbf{2.62} \\
				\cdashline{2-8}
				\specialrule{0em}{1pt}{1pt}
				& \textbf{CcGAN-AVAR-S} & \underline{0.048} & \textbf{1.728} & 3.705 & 1.273 & \underline{4162.04} & \underline{3.71} \\
				& \textbf{CcGAN-AVAR-H} & \textbf{0.042} & \underline{1.734} & \textbf{3.723} & \underline{1.270} & \underline{4162.04} & \underline{3.71} \\
				
				\midrule
				
				%% RC-49 (unimodal)
				\multirow{6}[1]{*}{\begin{tabular}[c]{@{}c@{}} \textbf{RC-49-I} \\ (unimodal) \\  {(64$\times$64)}\end{tabular}} &  CcDPM (AAAI'24) & 0.925 & 2.136 & 3.585 & 23.749  & 2.92 & 5.07 \\
				& CCDM (TMM'26) & 0.064 & 2.023  & 3.642 & \textbf{1.417} & 2.92 & 5.53 \\
				\cdashline{2-8}
				\specialrule{0em}{1pt}{1pt}
				& CcGAN (T-PAMI'23) & 0.151 & 1.880 & 3.400 & 2.487 & \textbf{6243.06} & \textbf{2.62} \\
				& Dual-NDA (AAAI'24) & 0.192 & 1.941 & 3.223  & 2.436  &  \textbf{6243.06} & \textbf{2.62} \\
				\cdashline{2-8}
				\specialrule{0em}{1pt}{1pt}
				& \textbf{CcGAN-AVAR-S} & \underline{0.054} & \textbf{1.729} & \underline{3.704} & 1.504 & \underline{4162.04} & \underline{3.71} \\
				&  \textbf{CcGAN-AVAR-H} & \textbf{0.050} & \underline{1.794} & \textbf{3.711} & \underline{1.425} & \underline{4162.04} & \underline{3.71} \\
				
				\midrule
				
				%% RC-49 (bimodal)
				\multirow{6}[1]{*}{\begin{tabular}[c]{@{}c@{}} \textbf{RC-49-I} \\ (bimodal) \\  {(64$\times$64)}\end{tabular}} & CcDPM (AAAI'24) & 0.850 & 2.214 & 3.572  & 21.972 & 2.92 & 5.07 \\
				& CCDM (TMM'26) & 0.066 & 2.012 & 3.659 & \textbf{1.272} & 2.92 & 5.53 \\
				\cdashline{2-8}
				\specialrule{0em}{1pt}{1pt}
				& CcGAN (T-PAMI'23) & 0.205 & 1.906 & 3.293 & 2.825 & \textbf{6243.06} & \textbf{2.62} \\
				& Dual-NDA (AAAI'24) & 0.235 & 1.886 & 3.162 & 2.317 & \textbf{6243.06} & \textbf{2.62} \\
				\cdashline{2-8}
				\specialrule{0em}{1pt}{1pt}
				& \textbf{CcGAN-AVAR-S} & \underline{0.052} & \underline{1.736} & \textbf{3.716} & 1.446 & \underline{4162.04} & \underline{3.71} \\
				& \textbf{CcGAN-AVAR-H} & \textbf{0.050} & \textbf{1.729} & \underline{3.706} & \underline{1.402} & \underline{4162.04} & \underline{3.71} \\
				
				\midrule
				
				%% RC-49 (trimodal)
				\multirow{6}[1]{*}{\begin{tabular}[c]{@{}c@{}} \textbf{RC-49-I} \\ (trimodal) \\  {(64$\times$64)}\end{tabular}} &  CcDPM (AAAI'24) & 0.940 & 2.224 & 3.551  & 23.902 & 2.92 & 5.07 \\
				& CCDM (TMM'26) & 0.060 & 2.047 & 3.664 & \textbf{1.117} & 2.92 & 5.53 \\
				\cdashline{2-8}
				\specialrule{0em}{1pt}{1pt}
				& CcGAN (T-PAMI'23) & 0.171 & 1.873 & 3.296 & 2.813 & \textbf{6243.06} & \textbf{2.62} \\
				& Dual-NDA (AAAI'24) & 0.168 & 1.949 & 3.302 & 2.575 & \textbf{6243.06} & \textbf{2.62} \\
				\cdashline{2-8}
				\specialrule{0em}{1pt}{1pt}
				& \textbf{CcGAN-AVAR-S} & \textbf{0.049} & \underline{1.735} & \textbf{3.716} & {1.490} & \underline{4162.04} & \underline{3.71} \\
				& \textbf{CcGAN-AVAR-H} & \underline{0.051} & \textbf{1.729} & \underline{3.707} & \underline{1.462} & \underline{4162.04} & \underline{3.71} \\
				
				\midrule
				
				%% UTKFace
				\multirow{6}[1]{*}{\begin{tabular}[c]{@{}c@{}} \textbf{UTKFace} \\ {(64$\times$64)}\end{tabular}} &  CcDPM (AAAI'24) & 0.466 & \underline{1.560} & 1.211 & 6.868  & 2.10 & 5.70 \\
				& CCDM (TMM'26) & 0.363 & \textbf{1.542} & 1.184 & \textbf{6.164} & 2.10 & 6.14 \\
				\cdashline{2-8}
				\specialrule{0em}{1pt}{1pt}
				& CcGAN (T-PAMI'23) & 0.413 & 1.733 & \textbf{1.329} & 8.240 & \textbf{5555.56} & \textbf{2.44} \\
				& Dual-NDA (AAAI'24) & 0.396 & 1.678 & \underline{1.298} & 6.765 & \textbf{5555.56} & \textbf{2.44} \\
				\cdashline{2-8}
				\specialrule{0em}{1pt}{1pt}
				& \textbf{CcGAN-AVAR-S} & \underline{0.358} & 1.703 & \underline{1.298} & 7.222 & \underline{4166.67} & \underline{3.71} \\		
				& \textbf{CcGAN-AVAR-H} & \textbf{0.356} & 1.691 & 1.278 & \underline{6.696} & \underline{4166.67} & \underline{3.71} \\
				
				\midrule
				
				%% Steering Angle
				\multirow{6}[1]{*}{\begin{tabular}[c]{@{}c@{}} \textbf{Steering}\\ \textbf{Angle} \\ {(64$\times$64)}\end{tabular}} &  CcDPM (AAAI'24) & 0.939  & \underline{1.761} & 1.150  &  10.999 & 3.33 & 2.28 \\
				& CCDM (TMM'26) & \textbf{0.742} & 1.778 & 1.088 & \textbf{5.823} & 3.33 & 2.74 \\
				\cdashline{2-8}
				\specialrule{0em}{1pt}{1pt}
				& CcGAN (T-PAMI'23) & 1.334 & 1.784 & \underline{1.234} & 14.807 & \textbf{3086.42} & \textbf{2.50} \\
				& Dual-NDA (AAAI'24) & 1.114 & \textbf{1.738} & \textbf{1.251} & 11.809 & \textbf{3086.42} & \textbf{2.50} \\
				\cdashline{2-8}
				\specialrule{0em}{1pt}{1pt}
				& \textbf{CcGAN-AVAR-S} & \underline{0.748} & 1.810 & 1.176 & \underline{6.408} & \underline{1157.41} & \underline{3.86} \\	
				& \textbf{CcGAN-AVAR-H} & 0.809 & 1.800  & 1.204   & {6.963} & \underline{1157.41} & \underline{3.86} \\	
				
				\bottomrule
			\end{tabular}%
			\end{adjustbox}
			
			\label{tab:main_results_low}%
		
	\end{table}

	%%%% Line graphs for candidate methods on RC49 (64x64)	
	
	%%%% Line graphs for candidate methods on UTKFace
	\begin{figure}[!htbp]
		\centering
		\includegraphics[width=0.9\linewidth]{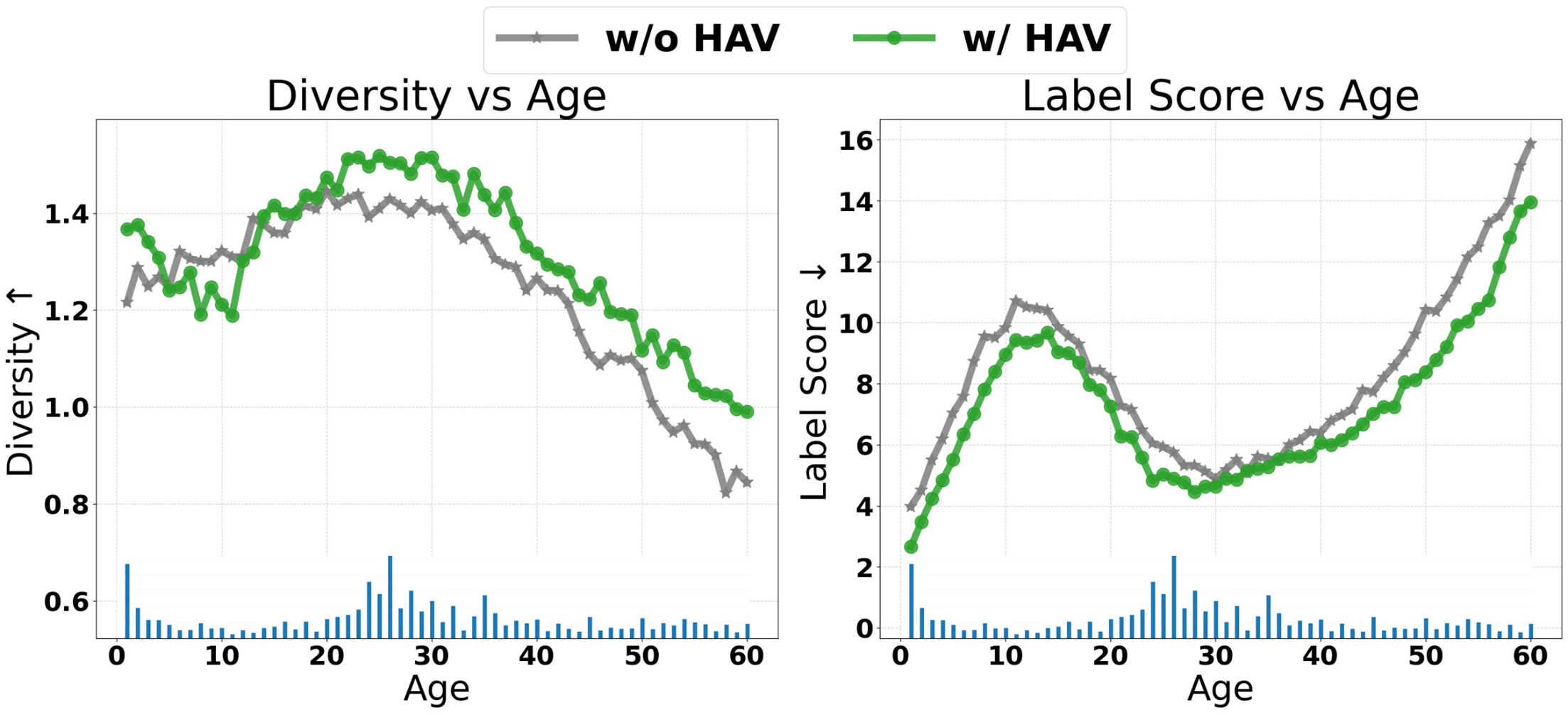} 
		\caption{ \textbf{Performance of the strong CcGAN baseline with and without the proposed Hybrid Adaptive Vicinity (HAV) on UTKFace ($64 \times 64$), with auxiliary generator regularization disabled.} The blue bar charts at the bottom of each image indicate sample sizes across different ages. HAV improves the overall diversity and label-consistency trends, although the gains vary across ages and minor local degradations may occur. } 
		\label{fig:line_graphs_uk64_wo_or_w_hav}
	\end{figure}
	
	%%%% Line graphs for sfid vs Nav
	\begin{figure}[!htbp]
		\centering
		\includegraphics[width=0.9\linewidth]{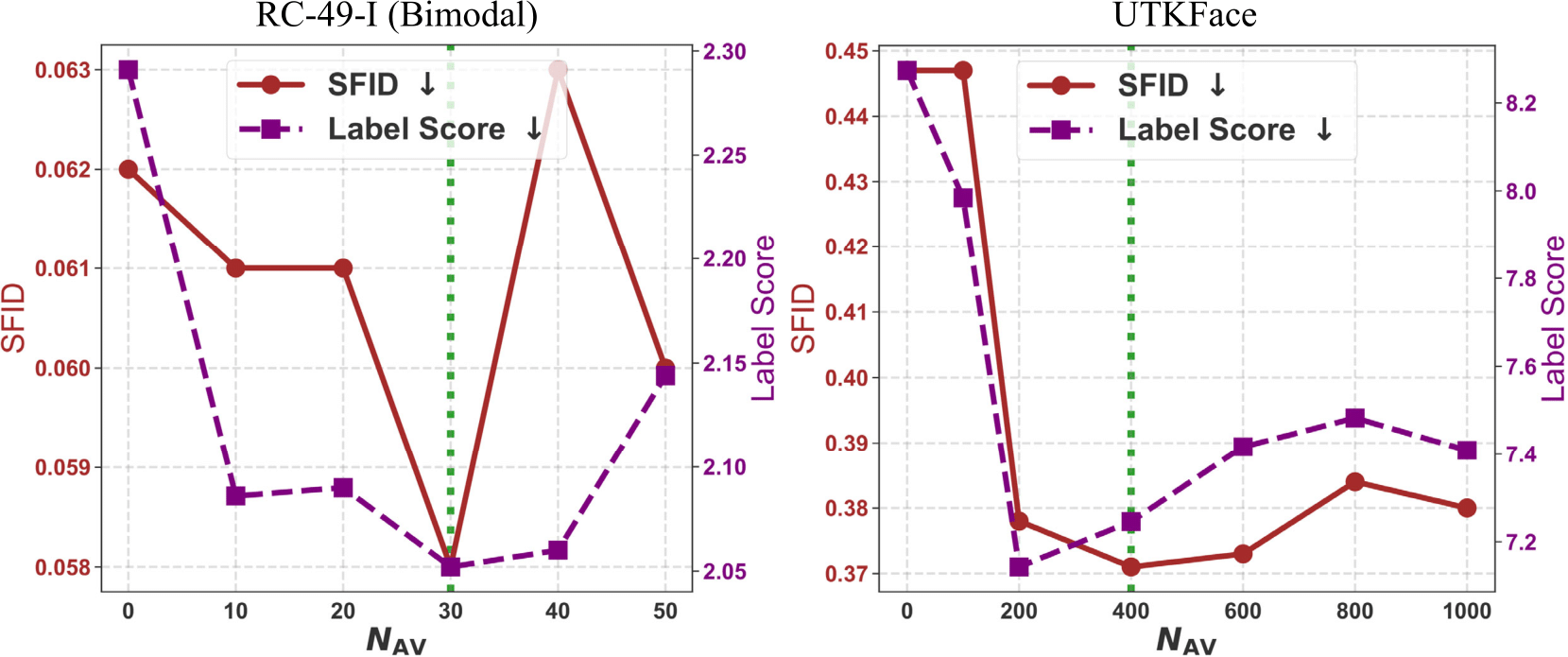} 
		\caption{ \textbf{The effect of the adaptive-vicinity sample-count threshold ($N_\text{AV}$) on CcGAN-AVAR-H's performance, comparing SFID and Label Score metrics across RC-49-I and UTKFace datasets at $64\times 64$ resolution.} Dashed green lines correspond to the values used in the primary experiment. } 
		\label{fig:ab_line_graph_sfid_vs_Nav}
	\end{figure}

	\begin{table}[!htbp]
        \setlength{\tabcolsep}{1mm} 
		\footnotesize
		\centering
			\caption{ \textbf{Comparison of generation quality and sampling efficiency at high resolutions ($>64\times 64$) across six compared methods.} Generation quality is evaluated by SFID (overall quality), NIQE (visual realism), Diversity (sample diversity), and Label Score (label consistency). Sampling efficiency is measured by inference speed and GPU memory consumption on a single NVIDIA RTX 4090D-48G GPU. ``$\downarrow$" (``$\uparrow$") indicates lower (higher) values are preferred. The best and second-best results are highlighted in bold and underlined, respectively. }
			\begin{adjustbox}{max totalsize={\textwidth}{1\textheight},center}
			\begin{tabular}{cc cccc cc}
				\toprule
				\begin{tabular}[c]{@{}c@{}} \textbf{Dataset} \\ (resolution)\end{tabular} & \textbf{Method} & \begin{tabular}[c]{@{}c@{}} \textbf{SFID} \\ {(primary)} \end{tabular} $\downarrow$   & \textbf{NIQE} $\downarrow$ & \textbf{Diversity} $\uparrow$ & \begin{tabular}[c]{@{}c@{}} \textbf{Label} \\ \textbf{Score} \end{tabular} $\downarrow$ & \begin{tabular}[c]{@{}c@{}} \textbf{Speed} \\ {(img/sec)}\end{tabular} $\uparrow$ & \begin{tabular}[c]{@{}c@{}} \textbf{Memory} \\ {(GiB)} \end{tabular} $\downarrow$ \\
				\midrule
				
				%% UTKFace (128x128)
				\multirow{6}[1]{*}{\begin{tabular}[c]{@{}c@{}} \textbf{UTKFace} \\ {(128$\times$128)}\end{tabular}} &  CcDPM (AAAI'24) & 0.529 & 1.114 & 1.195  & 7.933 & 1.12 & 17.91  \\
				& CCDM (TMM'26) & 0.319 & \textbf{1.077} & 1.178 & \underline{6.359} & 1.12 & 19.52 \\
				\cdashline{2-8}
				\specialrule{0em}{1pt}{1pt}
				& CcGAN (T-PAMI'23) & 0.367 & 1.113 & 1.199 & 7.747 & \textbf{1282.05} & \textbf{7.58}  \\
				& Dual-NDA (AAAI'24) & {0.361} & \underline{1.081} & \textbf{1.257} & \textbf{6.310} & \textbf{1282.05} & \textbf{7.58} \\
				\cdashline{2-8}
				\specialrule{0em}{1pt}{1pt}
				& \textbf{CcGAN-AVAR-S} & \underline{0.299} & 1.191 & 1.223 & 6.622 & \underline{1190.48} & \underline{11.57} \\		
				& \textbf{CcGAN-AVAR-H} & \textbf{0.297} & 1.173 & \underline{1.251} & 6.586 & \underline{1190.48} & \underline{11.57} \\
				
				\midrule
				
				%% Steering Angle (128x128)
				\multirow{6}[1]{*}{\begin{tabular}[c]{@{}c@{}} \textbf{Steering}\\ \textbf{Angle} \\ {(128$\times$128)}\end{tabular}} &  CcDPM (AAAI'24) & 1.285  & \underline{1.989}  & \textbf{1.203}   &  18.325  & 1.23 & \textbf{5.27} \\
				& CCDM (TMM'26) & 0.987 & \textbf{1.977} & 1.118 & 11.829 & 1.23 & \underline{6.85} \\
				\cdashline{2-8}
				\specialrule{0em}{1pt}{1pt}
				& CcGAN (T-PAMI'23) & {1.689} & {2.411} & {1.088} & {18.438} & \textbf{1543.21} & 7.58 \\
				& Dual-NDA (AAAI'24) & {1.390} & {2.135} & \underline{1.133} & {14.099} & \textbf{1543.21} & 7.58  \\
				\cdashline{2-8}
				\specialrule{0em}{1pt}{1pt}		
				& \textbf{CcGAN-AVAR-S} & \textbf{0.803} & 2.350 & 1.079 & \textbf{6.908} & \underline{1207.73} & 11.57 \\
				& \textbf{CcGAN-AVAR-H} & \underline{0.888} & 2.288 & 1.123 & \underline{7.507}  & \underline{1207.73} & 11.57 \\		
				
				\midrule
				
				%% UTKFace (192x192)
				\multirow{6}[1]{*}{\begin{tabular}[c]{@{}c@{}} \textbf{UTKFace} \\ {(192$\times$192)}\end{tabular}} &  CcDPM (AAAI'24) & 0.970  &  1.522  &  1.187  & 11.224  & 0.80 & 19.93 \\
				& CCDM (TMM'26) &  0.467 &   \textbf{1.242}    &  1.148 & 7.336 & 0.80 & 22.99 \\
				\cdashline{2-8}
				\specialrule{0em}{1pt}{1pt}
				& CcGAN (T-PAMI'23) & 0.499 & 1.661 & \textbf{1.207} & 7.885 & \underline{617.28}& \underline{18.18} \\
				& Dual-NDA (AAAI'24) & 0.487 & \underline{1.483} & \underline{1.201} & 6.730 & \underline{617.28} & \underline{18.18} \\
				\cdashline{2-8}
				\specialrule{0em}{1pt}{1pt}
				& \textbf{CcGAN-AVAR-S} & \underline{0.455} & 1.621 & 1.104 & \underline{6.433} & \textbf{1111.11} & \textbf{13.18} \\
				& \textbf{CcGAN-AVAR-H} & \textbf{0.435} & 1.588 & 1.187 & \textbf{6.377} & \textbf{1111.11} & \textbf{13.18} \\
				
				\midrule
				
				%% UTKFace (256x256)
				\multirow{6}[1]{*}{\begin{tabular}[c]{@{}c@{}} \textbf{UTKFace} \\ {(256$\times$256)}\end{tabular}} &  CcDPM (AAAI'24) & 0.559 & 1.552  & 0.854   & 12.121 & 0.30 & 18.17  \\
				& CCDM (TMM'26) & \underline{0.268} & 1.364 & 1.121 & 7.555 & 0.30 & 23.51  \\
				\cdashline{2-8}
				\specialrule{0em}{1pt}{1pt}
				& CcGAN (T-PAMI'23) & 0.347 & \textbf{1.308} & 1.053 & 6.010 & \underline{384.62} & \textbf{12.50}  \\
				& Dual-NDA (AAAI'24) & 0.319 & 1.381 & 1.216 & \underline{5.712} & \underline{384.62} & \textbf{12.50}  \\
				\cdashline{2-8}
				\specialrule{0em}{1pt}{1pt}
				& \textbf{CcGAN-AVAR-S} & \textbf{0.196} & 1.601 & \textbf{1.228} & \textbf{5.678} & \textbf{624.87} & \underline{16.39} \\
				& \textbf{CcGAN-AVAR-H} & \textbf{0.196}  & \underline{1.347} & \underline{1.219}  & 5.798 & \textbf{624.87} & \underline{16.39} \\
				
				\midrule
				
				%% Steering Angle (256x256)
				\multirow{6}[1]{*}{\begin{tabular}[c]{@{}c@{}} \textbf{Steering}\\ \textbf{Angle} \\ {(256$\times$256)}\end{tabular}} &  CcDPM (AAAI'24) & 1.217 & \underline{1.748} & \textbf{1.330} & 29.239 & 0.30 & 18.17  \\
				& CCDM (TMM'26) & 0.902 & \textbf{1.616} & \underline{1.151} & 23.057 & 0.30 & 23.31  \\
				\cdashline{2-8}
				\specialrule{0em}{1pt}{1pt}
				& CcGAN (T-PAMI'23) & 0.984 & 1.999 & 1.054 & 8.399 & \underline{613.50} & \underline{12.40}  \\
				& Dual-NDA (AAAI'24) & 0.967 & 1.834 & 0.951 & 8.338 & \underline{613.50} & \underline{12.40}  \\
				\cdashline{2-8}
				\specialrule{0em}{1pt}{1pt}		
				& \textbf{CcGAN-AVAR-S} & \underline{0.718} & 2.177 & 1.111 & \underline{6.522} & \textbf{626.17} & \textbf{11.97}  \\
				& \textbf{CcGAN-AVAR-H} & \textbf{0.683}  & 1.966  & 1.113  & \textbf{4.958} & \textbf{626.17} & \textbf{11.97}  \\
				
				\bottomrule
			\end{tabular}%
			\end{adjustbox}
			
			\label{tab:main_results_high}%
		
	\end{table}%

	%------------------------------------------------
	\subsection{Ablation Study}\label{sec:ablation}

	We perform ablation studies on RC-49-I and UTKFace at $64\times 64$ and $128\times 128$ resolutions to systematically assess the impact of key components and hyperparameters in CcGAN-AVAR. These experiments complement the theoretical interpretation in Section~\ref{sec:theory_interpretation} by empirically examining the effects of adaptive vicinity, auxiliary regularization, and the radius-construction threshold $N_\text{AV}$, as detailed below:
	\begin{itemize}
		
		\item \textbf{The Effects of AV and Regularization}: Our component-wise analysis of CcGAN-AVAR on RC-49-I and UTKFace (Table \ref{tab:ablation_components}) reveals that the strong CcGAN baseline, implemented using the unified codebase with SNGAN backbone, exponential moving average (EMA), and mixed-precision training, achieves comparable or superior generation performance to the vanilla CcGAN \cite{ding2023ccgan} marked with an asterisk (*) while requiring substantially less training cost. By incrementally incorporating the proposed soft/hybrid adaptive vicinity, regression penalty, and $f$-divergence penalty into this baseline, the generation performance is progressively improved. 
		
		% We also evaluate the effect of HAV by comparing the strong CcGAN baseline with and without it on the UTKFace dataset ($64\times 64$), as shown in Fig.~\ref{fig:line_graphs_uk64_wo_or_w_hav}. In this ablation study, auxiliary generator regularization was disabled to isolate HAV's contribution. The results demonstrate that HAV consistently improves both sample diversity and label consistency. The bar chart at the bottom displays the sample size for each age.
        We also evaluate the effect of HAV by comparing the strong CcGAN baseline with and without it on the UTKFace dataset ($64\times64$), as shown in Fig.~\ref{fig:line_graphs_uk64_wo_or_w_hav}. In this ablation study, auxiliary generator regularization was disabled to isolate HAV's contribution. In this controlled comparison, HAV improves the overall diversity and label consistency trends, although the gains can vary across ages. The bar chart at the bottom displays the sample size for each age.
		
		\item \textbf{The Effect of Network Architecture}: We evaluate the generation performance of CcGAN-AVAR-H using different network architectures, including DCGAN~\cite{radford2015unsupervised}, SNGAN~\cite{miyato2018spectral}, SAGAN~\cite{zhang2019self}, and BigGAN (including BigGAN-deep)~\cite{brock2018large}. The quantitative results, summarized in Table \ref{tab:ab_effect_network}, show that the SNGAN backbone performs best on CCGM benchmark datasets. This outcome validates the network design we proposed in Section \ref{sec:network_arch}. 
		
		\item \textbf{The Effects of $N_\text{AV}$}: We investigate the impact of the key hyperparameter $N_\text{AV}$ (the adaptive-vicinity sample-count threshold) on CcGAN-AVAR's performance (Fig.~\ref{fig:ab_line_graph_sfid_vs_Nav}), conducting this ablation study with auxiliary regularization disabled to isolate AV's effect. The results demonstrate substantial performance degradation when either disabling AV entirely ($N_\text{AV}=0$) or employing an excessively wide vicinity (large $N_\text{AV}$ values), while moderate $N_\text{AV}$ values consistently yield acceptable results. The dashed green lines in Fig.~\ref{fig:ab_line_graph_sfid_vs_Nav} correspond to the optimal values selected for our main experiments.
		
	\end{itemize}

	\begin{table}[!htbp]
		\setlength{\tabcolsep}{1mm} 
		\footnotesize
		\centering
		\caption{\textbf{Ablation study: The impact of network architecture on CcGAN-AVAR-H.} Since more complex architectures such as SAGAN, BigGAN, BigGAN-deep underperform SNGAN, we adopt SNGAN as the backbone for our CcGAN-AVAR implementation.}
			   % \begin{adjustbox}{width=1\linewidth}
				\begin{tabular}{cccccc}
					\toprule
					\textbf{Dataset} & \begin{tabular}[c]{@{}c@{}} \textbf{Network} \\ \textbf{Architecture}  \end{tabular}    & \begin{tabular}[c]{@{}c@{}} \textbf{SFID} \\ {(primary)} \end{tabular} & \textbf{NIQE} & \textbf{Diversity} & \begin{tabular}[c]{@{}c@{}} \textbf{Label} \\ \textbf{Score} \end{tabular} \\
					\midrule
					\multirow{4}[1]{*}{\begin{tabular}[c]{@{}c@{}} \textbf{RC-49-I} \\ (bimodal) \\ ($64\times 64$) \end{tabular}} & SNGAN   & 0.050 & 1.729 & 3.706 & 1.402 \\
					& DCGAN   & 0.063 & 1.737  & 3.636  & 1.585  \\
					& SAGAN   & 0.082 & 1.972  & 3.631  & 1.360  \\
					& BigGAN   & 0.071  & 1.958 & 3.650  & 1.458  \\
					%				& BigGAN-deep   & 0.447 & 2.073  & 2.563  & 4.935  \\
					\cdashline{1-6}
					\specialrule{0em}{1pt}{1pt}
					
					\multirow{4}[0]{*}{\begin{tabular}[c]{@{}c@{}} \begin{tabular}[c]{@{}c@{}} \textbf{UTKFace} \\ ($64\times 64$) \end{tabular} \end{tabular}} & SNGAN & 0.356 & 1.691 & 1.278 & 6.696 \\
					& DCGAN   & 0.363  & 1.690  & 1.262  & 6.701  \\
					& SAGAN   & 0.396  & 1.657  & 1.228  & 7.159  \\
					& BigGAN   & 0.369  & 1.741  & 1.261  & 7.241  \\
					%				& BigGAN-deep   & 0.436  & 1.730  & 1.224  & 6.684  \\
					
					\cdashline{1-6}
					\specialrule{0em}{1pt}{1pt}
					
					\multirow{4}[0]{*}{\begin{tabular}[c]{@{}c@{}} \begin{tabular}[c]{@{}c@{}} \textbf{UTKFace} \\ ($128\times 128$) \end{tabular} \end{tabular}} & SNGAN   & 0.297 & 1.173 & 1.251 & 6.586 \\		
					& SAGAN   & 0.378 & 1.255  & 1.183 & 6.759 \\
					& BigGAN   & 0.337 & 1.130 & 1.215 & 7.464 \\
					& BigGAN-deep   & 0.464  & 1.248  & 1.173  & 4.837 \\
					
					\bottomrule
				\end{tabular}%
				% \end{adjustbox}
			\label{tab:ab_effect_network}%
	\end{table}%

	% Component-wise ablation study
	\begin{table}[!htbp]
		\setlength{\tabcolsep}{1mm} 
		\footnotesize
		\centering
			\caption{\textbf{Ablation study: Component-wise analysis of CcGAN-AVAR.} The strong CcGAN achieves comparable or superior performance to the vanilla CcGAN \cite{ding2023ccgan} marked with an asterisk (*) while requiring substantially less training cost. Performance progressively improves by incrementally incorporating: (1) soft/hybrid adaptive vicinity, (2) regression penalty $\mathcal{L}^G_\text{reg}$, and (3) $f$-divergence penalty $\mathcal{L}^G_f$. }
			% \begin{adjustbox}{width=1\linewidth}			
			\begin{tabular}{c ccccc cccc}
				\toprule
				\multirow{3}[0]{*}{\textbf{Dataset}}  & \multicolumn{5}{c}{\textbf{Configuration}} & \multirow{3}[0]{*}{\textbf{SFID}} & \multirow{3}[0]{*}{\textbf{NIQE}} & \multirow{3}[0]{*}{\textbf{Diversity}} & \multirow{3}[0]{*}{\begin{tabular}[c]{@{}c@{}} \textbf{Label} \\ \textbf{Score} \end{tabular}} \\
				\textbf{} &  \begin{tabular}[c]{@{}c@{}} \textbf{Strong} \\ \textbf{CcGAN} \end{tabular} &  \textbf{SAV} &  \textbf{HAV} & $\mathcal{L}^G_\text{reg}$ & $\mathcal{L}^G_f$ & \textbf{} & \textbf{} & \textbf{} & \textbf{} \\
				\midrule
				\multirow{5}[0]{*}{\begin{tabular}[c]{@{}c@{}} \textbf{RC-49-I} \\ (bimodal) \\ ($64\times 64$) \end{tabular}} &    &    &    &   &   & 0.205* & 1.906* & 3.293* & 2.825* \\
				&  \checkmark &    &    &   &   & 0.062 & 1.740 & 3.704 & 2.291 \\
				&  \checkmark& \checkmark &  &  &  & 0.059 & 1.732 & 3.700 & 2.238 \\
				&  \checkmark& \checkmark &  & \checkmark &   & 0.055 & 1.727 & 3.695 & 1.737 \\
				&  \checkmark& \checkmark &  & \checkmark & \checkmark & 0.052 & 1.736 & 3.716 & 1.446 \\
				\midrule
				\multirow{5}[0]{*}{\begin{tabular}[c]{@{}c@{}} \textbf{RC-49-I} \\ (bimodal) \\ ($64\times 64$) \end{tabular}}&    &      &    &   & & 0.205* & 1.906* & 3.293* & 2.825* \\
				&   \checkmark &      &    &   & & 0.062 & 1.740 & 3.704 & 2.291 \\
				&   \checkmark& & \checkmark &  &  & 0.058 & 1.739 & 3.708 & 2.052 \\
				&   \checkmark& & \checkmark   & \checkmark &   & 0.050 & 1.723 & 3.715 & 1.644 \\
				&   \checkmark& & \checkmark   & \checkmark & \checkmark & 0.050 & 1.729 & 3.706 & 1.402 \\
				\midrule
				\multirow{5}[0]{*}{\begin{tabular}[c]{@{}c@{}} \textbf{UTKFace} \\ ($64\times 64$) \end{tabular}} &   &    &     &  &   & 0.413* & 1.733* & 1.329* & 8.240* \\
				&   \checkmark &    &     &  &   & 0.447 & 1.738 & 1.243 & 8.274 \\
				&   \checkmark&  \checkmark & &     &     & 0.442 & 1.750 & 1.305 & 7.867 \\
				&   \checkmark&  \checkmark & & \checkmark   &     & 0.432 & 1.747 & 1.307 & 8.073 \\
				&   \checkmark&  \checkmark & & \checkmark   & \checkmark   & 0.358 & 1.703 & 1.298 & 7.222 \\
				\midrule
				\multirow{5}[0]{*}{\begin{tabular}[c]{@{}c@{}} \textbf{UTKFace} \\ ($64\times 64$) \end{tabular}} &   &   &   &     &     & 0.413* & 1.733* & 1.329* & 8.240* \\
				&   \checkmark &   &   &     &     & 0.447 & 1.738 & 1.243 & 8.274 \\
				&   \checkmark& & \checkmark  &     &     & 0.371 & 1.696 & 1.310 & 7.246 \\
				&   \checkmark& &  \checkmark & \checkmark   &     & 0.357 & 1.709 & 1.269 & 6.529 \\
				&   \checkmark& &  \checkmark & \checkmark   & \checkmark   & 0.356 & 1.691 & 1.278 & 6.696 \\
				\bottomrule
			\end{tabular}%
			% \end{adjustbox}
			\label{tab:ablation_components}%
	\end{table}%

	%%%%%%%%%%%%%%%%%%%%%%%%%%%%%%%%%%%%%%%%%%%%%%%%%%%%%%%%%%%%%%%%%%%
	\section{Conclusion}
	\label{sec:conclusion}
	
    This paper presents CcGAN-AVAR, an imbalance-aware extension of CcGAN for continuous conditional image generation. The proposed framework introduces soft and hybrid adaptive vicinity mechanisms to replace conventional fixed-size vicinities, together with a dual regularization strategy that incorporates auxiliary regression and density-ratio-based $f$-divergence penalties during generator training. In addition, we provide a theoretical interpretation showing how adaptive vicinal weighting affects the estimator-level local bias--variance behavior, how HAV reduces objective-level cross-condition mixing, and how the DRE-based generator penalty approximates a Pearson $\chi^2$ discrepancy under density-ratio estimation error. Extensive experiments across eleven settings with varying resolutions demonstrate that CcGAN-AVAR provides strong generation quality and label consistency while preserving the one-step sampling efficiency of GANs. 

    %%%%%%%%%%%%%%%%%%%%%%%%%%%%%%%%%%%%%%%%%%%%%%%%%%%%%%%%%%%%%%%%%%%
    \section*{Acknowledgments}
    This work was supported in part by the National Natural Science Foundation of China (Grant No.~62306147, U24A20326, 62502442, 62306144), Basic Research Program of Jiangsu (Grant No.~BK20241901, BK20230379), Jiangsu Supplementary Funds of National Major Talent Project (2022--2024) (Grant No.~R2024PT03), and the Startup Foundation for Introducing Talent of NUIST (Grant No.~2024r056).

% \clearpage
\bibliographystyle{elsarticle-num}
\bibliography{bibliography}

\clearpage
\appendix
\setcounter{equation}{0}
\renewcommand{\theequation}{S.\arabic{equation}}
\numberwithin{figure}{section}
\numberwithin{table}{section}

% Extracted from appendix.tex for inclusion in the Neurocomputing manuscript.
\section{GitHub repository}\label{supp:codes}
The source code and implementation details are publicly available at:
\begin{center}
	\url{https://github.com/UBCDingXin/CcGAN-AVAR}
\end{center}

%%%%%%%%%%%%%%%%%%%%%%%%%%%%%%%%%%%%%%%%%%%%%%%%%%%%%%%%%%%%%%%%%%%
\section{Experimental Setup and Implementation}

\subsection{Construction of RC-49-I}

\begin{figure}[H]  %[!htbp] 
	\centering
	\includegraphics[width=1\linewidth]{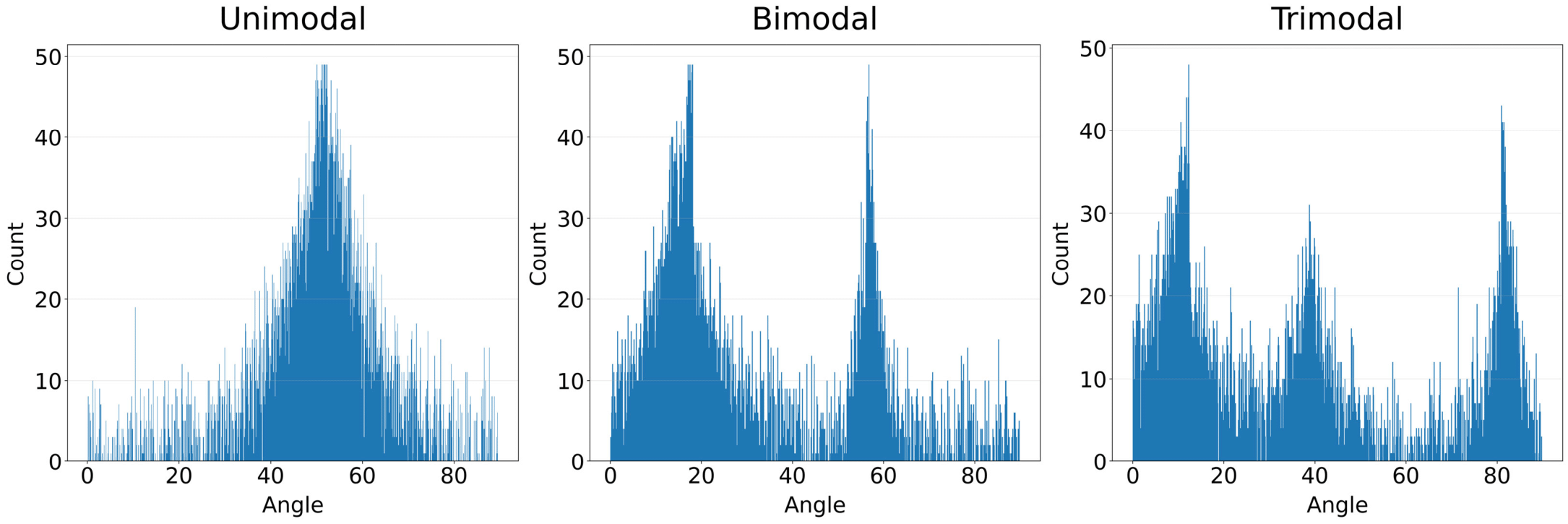}
	\caption{Three data imbalance patterns in RC-49-I}
	\label{fig:supp_label_dist_rc64_imb}
\end{figure}

Since RC-49 is a relatively simple synthetic dataset with a sufficiently large sample size for reliable evaluation, we construct three imbalanced subsets to thoroughly assess the robustness of CcGAN-AVAR and other baseline methods. As shown in Fig. \ref{fig:supp_label_dist_rc64_imb}, these subsets exhibit unimodal, bimodal, and trimodal label distributions. Using the unimodal case as an example, we describe the derivation procedure from RC-49 as follows: 
\begin{itemize}
	\item First, we randomly select a distinct label $y_{m} \in Y^u$ as the mode of the imbalanced distribution.
	
	\item Second, for each element $y_{(i)}$ in $Y^u$ excluding $y_{m}$, compute its absolute distance to $y_{m}$ as $d_i=|y_{(i)}-y_{m}|$. 
	
	\item Third, determine the mean sample count for $y_{(i)}$ using $\bar{N}^{\circ}_i=\max(1, \operatorname{int}(49\times e^{-\pi\cdot d_i}))$, where $\pi$ is a decay rate (set to 0.1 for unimodal distributions). The final sample count for $y_{(i)}$ is then calculated with noise perturbation: $N^{\circ}_i=\min(49, \max(0, \operatorname{int}(\bar{N}^{\circ}_i+\mathcal{N}(0,5)) ))$
	
	\item Finally, we construct RC-49-I (unimodal) in a label-by-label manner by randomly selecting $N^{\circ}_i$ images from the complete set of 49 images for each $y_{(i)}$.
	
\end{itemize}
The bimodal and trimodal subsets require a more sophisticated generation procedure, with full implementation details available in our codebase.

\subsection{Construction of the $256\times 256$ versions of UTKFace and Steering Angle}

{\setlength{\parindent}{0cm}\textbf{UTKFace.}} We perform super-resolution on the $192\times 192$ version of UTKFace using pre-trained Real-ESRGAN \cite{wang2021real} and GFPGAN \cite{wang2021towards}. First, Real-ESRGAN\footnote{\url{https://github.com/xinntao/Real-ESRGAN/releases/download/v0.1.0/RealESRGAN_x4plus.pth}} is employed to reconstruct $256\times 256$ images from the $192\times 192$ inputs, primarily enhancing background textures and overall image quality. Subsequently, GFPGAN\footnote{\url{https://github.com/TencentARC/GFPGAN/releases/download/v1.3.0/GFPGANv1.3.pth}} detects facial regions in the super-resolved images. After alignment, these regions are processed by a face restoration network. The restoration results are then inversely transformed to their original positions via an affine transformation and blended with the background using mask-based weighted fusion, with boundary feathering applied to minimize artifacts. Finally, the enhanced image is resampled to the target resolution and saved in a new HDF5 file, preserving the original data format. 

{\setlength{\parindent}{0cm}\textbf{Steering Angle.}} We apply the same pre-trained Real-ESRGAN to perform super-resolution on the $128\times 128$ Steering Angle dataset, saving the reconstructed images to a new HDF5 file.

\begin{figure}[H]  %[!htbp] 
	\centering
	\includegraphics[width=0.85\linewidth]{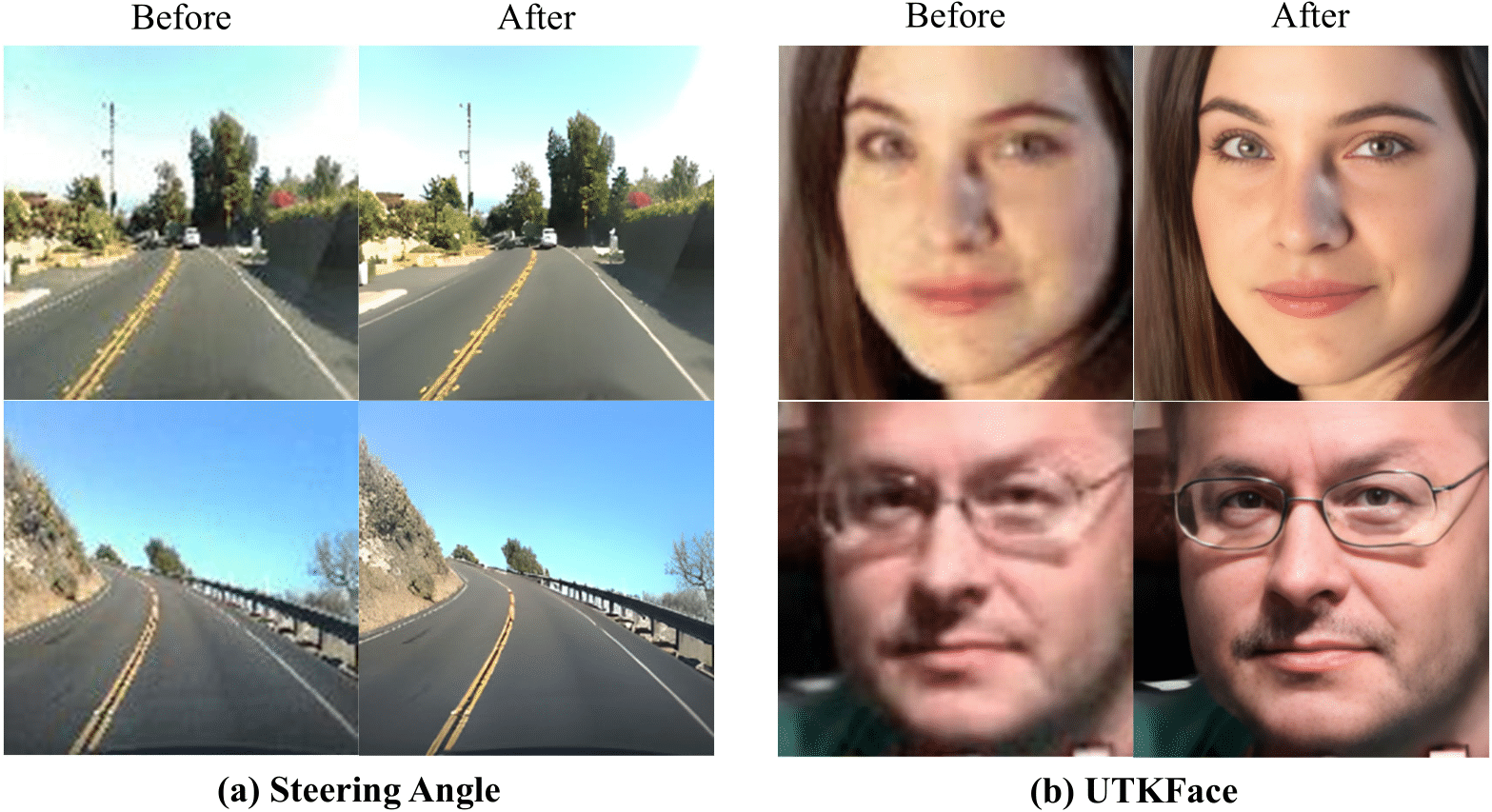}
	\caption{Example images of UTKFace and Steering Angle before and after super-resolution.}
	\label{fig:UK256_and_SA256_SR}
\end{figure}

Fig.~\ref{fig:UK256_and_SA256_SR} shows some example images from both datasets before and after super-resolution. The datasets are available for download via the links provided in our code repository.

\subsection{Hardware and Software Details}

All experiments were conducted on NVIDIA RTX 4090D servers with the following hardware configurations:
\begin{itemize}
	\item $64\times 64$ resolution: RTX 4090D-24G
	\item $>64\times 64$ resolutions: RTX 4090D-48G (modified)
\end{itemize}

The key software requirements include:
\begin{itemize}
	\item \textbf{Operating System:} Ubuntu 22.04 LTS
	\item \textbf{GPU Computing:} CUDA 12.8
	\item \textbf{Core Software:} Python 3.12.7; MATLAB R2021b
	\item \textbf{Python Packages:} torch 2.7.0; torchvision 0.22.0; accelerate 1.6.0; numpy 1.26.4; scipy 1.13.1; h5py 3.11.0; Pillow 10.4.0; matplotlib 3.9.2
\end{itemize}

\subsection{Training Configuration Details}

For CcGAN-AVAR implementation, we train a ResNet-18 model from scratch on each dataset's training set to predict image regression labels, using 200 epochs with a batch size of 256, base learning rate of 0.01, weight decay of $10^{-4}$, and enabled mixed-precision training.

Table \ref{tab:supp_train_cofigs} details our CcGAN-AVAR training configurations, which employ a modified SNGAN backbone with exponential moving average and mixed-precision training---improvements over \cite{ding2023ccgan} and \cite{ding2024turning}. Through mixed-precision training, our implementation demonstrates a $2\times$ improvement in training speed and 50\% reduction in memory usage compared to \cite{ding2023ccgan} when using identical architecture and batch size configurations.

\begin{table}[H]
	\small
	\centering
	\caption{Training Configurations for CcGAN-AVAR}
	\begin{adjustbox}{max width=\textwidth}
		\begin{tabular}{cl}
			\toprule
			\textbf{Dataset} & \textbf{Configuration} \\
			\midrule
			\textbf{RC-49} & \begin{tabular}[c]{@{}l@{}} SNGAN, gene\_ch=64, disc\_ch=48, hinge loss, DiffAugment, use EMA, steps=50K, batch size=256, lr=$10^{-4}$, \\ num\_D\_steps=2, $N_\text{AV}=50$, $\lambda_\text{dre}=10^{-2}$, $\lambda_\text{reg}^D=\lambda_\text{reg}^G=1$, $\lambda_\text{dre}^D=1$, $\lambda_f^G=0.5$ \end{tabular} \\
			\midrule
			\begin{tabular}[c]{@{}c@{}} \textbf{RC-49-I} \\ (unimodal) \end{tabular} & \begin{tabular}[c]{@{}l@{}} SNGAN, gene\_ch=64, disc\_ch=48, hinge loss, DiffAugment, use EMA, steps=50K, batch size=256, lr=$10^{-4}$, \\ num\_D\_steps=2, $N_\text{AV}=30$, $\lambda_\text{dre}=10^{-2}$, $\lambda_\text{reg}^D=\lambda_\text{reg}^G=1$, $\lambda_\text{dre}^D=1$ (hybrid), $\lambda_\text{dre}^D=0.5$ (soft), $\lambda_f^G=0.5$ \end{tabular} \\
			\midrule
			\begin{tabular}[c]{@{}c@{}} \textbf{RC-49-I} \\ (bimodal) \end{tabular} & \begin{tabular}[c]{@{}l@{}} SNGAN, gene\_ch=64, disc\_ch=48, hinge loss, DiffAugment, use EMA, steps=50K, batch size=256, lr=$10^{-4}$, \\ num\_D\_steps=2, $N_\text{AV}=30$, $\lambda_\text{dre}=10^{-2}$, $\lambda_\text{reg}^D=\lambda_\text{reg}^G=1$, $\lambda_\text{dre}^D=1$ (hybrid), $\lambda_\text{dre}^D=0.5$ (soft), $\lambda_f^G=0.5$ \end{tabular} \\
			\midrule
			\begin{tabular}[c]{@{}c@{}} \textbf{RC-49-I} \\ (trimodal) \end{tabular} & \begin{tabular}[c]{@{}l@{}} SNGAN, gene\_ch=64, disc\_ch=48, hinge loss, DiffAugment, use EMA, steps=50K, batch size=256, lr=$10^{-4}$, \\ num\_D\_steps=2, $N_\text{AV}=30$, $\lambda_\text{dre}=10^{-2}$, $\lambda_\text{reg}^D=\lambda_\text{reg}^G=1$, $\lambda_\text{dre}^D=1$ (hybrid), $\lambda_\text{dre}^D=0.5$ (soft), $\lambda_f^G=0.5$ \end{tabular} \\
			\midrule
			\begin{tabular}[c]{@{}c@{}} \textbf{UTKFace} \\ (64$\times$64) \end{tabular} & \begin{tabular}[c]{@{}l@{}} SNGAN, gene\_ch=64, disc\_ch=48, hinge loss, DiffAugment, use EMA, steps=50K, batch size=256, lr=$10^{-4}$, \\ num\_D\_steps=2, $N_\text{AV}=400$, $\lambda_\text{dre}=10^{-2}$, $\lambda_\text{reg}^D=\lambda_\text{reg}^G=1$, $\lambda_\text{dre}^D=0.5$, $\lambda_f^G=0.5$ \end{tabular} \\
			\midrule
			\begin{tabular}[c]{@{}c@{}} \textbf{Steering Angle} \\ (64$\times$64) \end{tabular} & \begin{tabular}[c]{@{}l@{}} SNGAN, gene\_ch=64, disc\_ch=64, hinge loss, DiffAugment, use EMA, steps=200K (hybrid) / 190K (soft), \\ batch size=256, lr=$10^{-4}$, num\_D\_steps=1, $N_\text{AV}=20$, $\lambda_\text{dre}=10^{-2}$, $\lambda_\text{reg}^D=\lambda_\text{reg}^G=1$, $\lambda_\text{dre}^D=0.5$, $\lambda_f^G=0.5$ \end{tabular} \\
			\midrule
			\begin{tabular}[c]{@{}c@{}} \textbf{UTKFace} \\ (128$\times$128) \end{tabular} & \begin{tabular}[c]{@{}l@{}} SNGAN, gene\_ch=64, disc\_ch=48, hinge loss, DiffAugment, use EMA, steps=60K, batch size=128, lr=$10^{-4}$, \\ num\_D\_steps=2, $N_\text{AV}=400$, $\lambda_\text{dre}=10^{-2}$, $\lambda_\text{reg}^D=\lambda_\text{reg}^G=1$, $\lambda_\text{dre}^D=0.5$, $\lambda_f^G=0.5$ \end{tabular} \\
			\midrule
			\begin{tabular}[c]{@{}c@{}} \textbf{Steering Angle} \\ (128$\times$128) \end{tabular} & \begin{tabular}[c]{@{}l@{}} SNGAN, gene\_ch=64, disc\_ch=48, hinge loss, DiffAugment, use EMA, steps=200K (hybrid) / 190K (soft), \\ batch size=128, lr=$10^{-4}$, num\_D\_steps=1, $N_\text{AV}=20$, $\lambda_\text{dre}=10^{-2}$, $\lambda_\text{reg}^D=\lambda_\text{reg}^G=1$, $\lambda_\text{dre}^D=0.5$, $\lambda_f^G=0.5$ \end{tabular} \\
			\midrule
			\begin{tabular}[c]{@{}c@{}} \textbf{UTKFace} \\ (192$\times$192) \end{tabular} & \begin{tabular}[c]{@{}l@{}} SNGAN, gene\_ch=64, disc\_ch=48, hinge loss, DiffAugment, use EMA, steps=70K (hybrid) / 50K (soft), \\ batch size=128, lr=$10^{-4}$, num\_D\_steps=2, $N_\text{AV}=400$, $\lambda_\text{dre}=10^{-2}$, $\lambda_\text{reg}^D=\lambda_\text{reg}^G=1$, $\lambda_\text{dre}^D=0.5$, $\lambda_f^G=0.5$, \\ DRE branch uses Dropout \end{tabular} \\
			\midrule
			\begin{tabular}[c]{@{}c@{}} \textbf{UTKFace} \\ (256$\times$256) \end{tabular} & \begin{tabular}[c]{@{}l@{}} SNGAN, gene\_ch=64, disc\_ch=48, hinge loss, DiffAugment, use EMA, steps=100K, batch size=64, \\ num\_grad\_acc=2, lr=$10^{-4}$, num\_D\_steps=2, $N_\text{AV}=400$, $\lambda_\text{dre}=10^{-2}$, $\lambda_\text{reg}^D=\lambda_\text{reg}^G=1$, $\lambda_\text{dre}^D=0.5$, $\lambda_f^G=0.5$ \end{tabular} \\
			\midrule
			\begin{tabular}[c]{@{}c@{}} \textbf{Steering Angle} \\ (256$\times$256) \end{tabular} & \begin{tabular}[c]{@{}l@{}} SNGAN, gene\_ch=64, disc\_ch=48, hinge loss, DiffAugment, use EMA, steps=80K (hybrid) / 30K (soft), \\ batch size=64, num\_grad\_acc=2, lr=$10^{-4}$, num\_D\_steps= 1 (hybrid) / 2 (soft), $N_\text{AV}=20$, $\lambda_\text{dre}=10^{-2}$, \\ $\lambda_\text{reg}^D=\lambda_\text{reg}^G=1$, 
				$\lambda_\text{dre}^D=0.5$, $\lambda_f^G=0.5$ \end{tabular} \\
			\bottomrule
		\end{tabular}%
	\end{adjustbox}
	\label{tab:supp_train_cofigs}%
\end{table}%

\subsection{Evaluation Configuration Details}

Our evaluation protocol follows \cite{ding2023ccgan, ding2024turning, ding2025ccdm}, with detailed specifications provided in Appendix S.V.B of \cite{ding2025ccdm}. For the RC-49-I experiments, we maintain identical evaluation procedures to those used for RC-49. 

The sampling batch size was set to 200 for resolutions below $256\times 256$ and 100 for $256\times 256$ experiments.

%%%%%%%%%%%%%%%%%%%%%%%%%%%%%%%%%%%%%%%%%%%%%%%%%%%%%%%%%%%%%%%%%%%
\section{Extra Experimental Results}

\begin{figure}[H]
	\centering
	\includegraphics[width=1\linewidth]{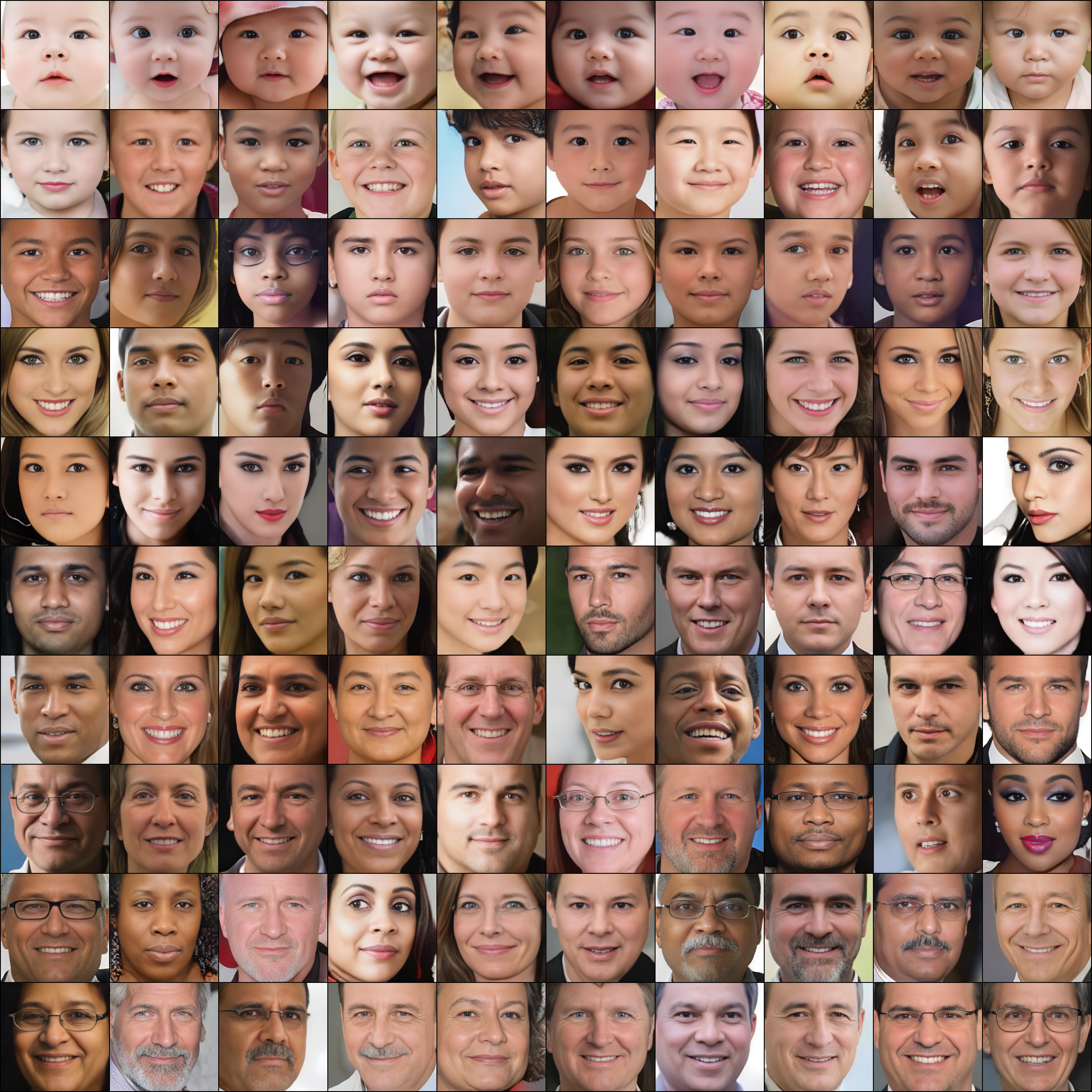}
	\caption{Fake images generated by CcGAN-AVAR-H for the UTKFace dataset at 256$\times$256 resolution. Each row corresponds to a specific age.}\label{fig:UK256_example_fake_images}
\end{figure}

\begin{figure}[H]
	\centering
	\includegraphics[width=1\linewidth]{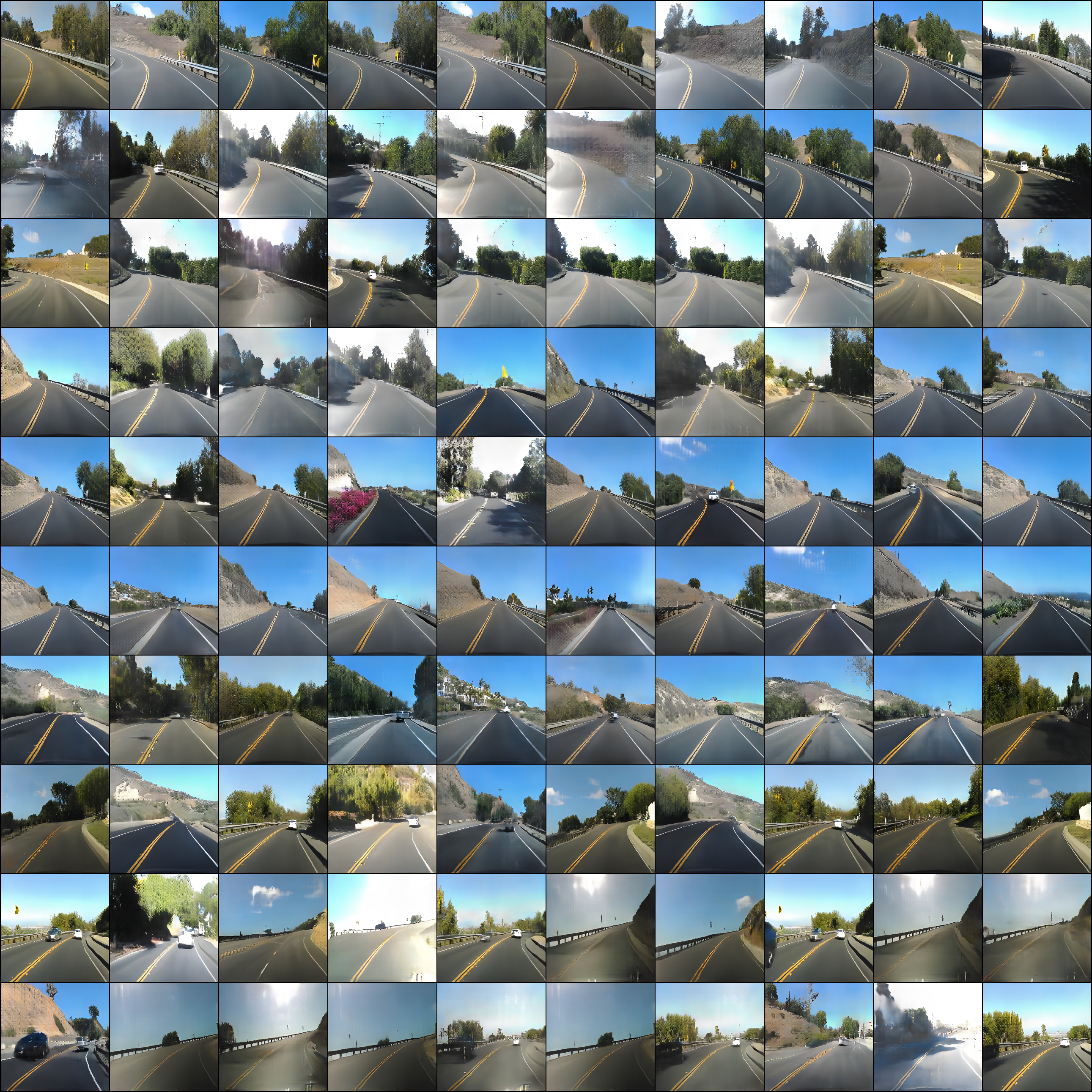}
	\caption{Fake images generated by CcGAN-AVAR-H for the Steering Angle dataset at 256$\times$256 resolution. Each row corresponds to a specific angle.}\label{fig:SA256_example_fake_images}
\end{figure}

Example synthetic images generated by CcGAN-AVAR-H for the UTKFace and Steering Angle experiments (256$\times$256 resolution) are presented in Figs.~\ref{fig:UK256_example_fake_images} and \ref{fig:SA256_example_fake_images}.

%% Example UK128 fake images

%% Example SA128 fake images

%%%%%%%%%%%%%%%%%%%%%%%%%%%%%%%%%%%%%%%%%%%%%%%%%%%%%%%%%%%%%%%%%%%
\section{Proofs for the Theoretical Interpretation}
\label{supp:proofs}

This section provides the supplementary proofs for the propositions introduced in Section~\ref{sec:theory_interpretation} of the main paper. The proofs are intended to clarify the objective-level interpretation of adaptive vicinal training and the density-ratio-estimation-based generator penalty.

\noindent\textbf{Proof of Proposition~\ref{prop:vicinal_bias}.}
For any $f\in\mathcal{F}$, expanding the expectation under the vicinal mixture gives
\begin{align}
	&\left|\mathbb{E}_{P_W}f(\bm{x})-\mathbb{E}_{P_{y_c}}f(\bm{x})\right| \nonumber\\
	&=\left|\sum_i W_i(y_c)\left(\mathbb{E}_{P_{y_i}}f(\bm{x})-\mathbb{E}_{P_{y_c}}f(\bm{x})\right)\right| \nonumber\\
	&\leq \sum_i W_i(y_c)\left|\mathbb{E}_{P_{y_i}}f(\bm{x})-\mathbb{E}_{P_{y_c}}f(\bm{x})\right|.
\end{align}
The inequality follows from the triangle inequality and the non-negativity of the weights. By the definition of the integral probability metric, we have
\begin{equation}
	\left|\mathbb{E}_{P_{y_i}}f(\bm{x})-\mathbb{E}_{P_{y_c}}f(\bm{x})\right|
	\leq d_{\mathcal{F}}(P_{y_i},P_{y_c}).
\end{equation}
Combining this bound with the assumed local Lipschitz condition, $d_{\mathcal{F}}(P_y,P_{y_c})\leq L|y-y_c|$, yields
\begin{equation}
	\left|\mathbb{E}_{P_W}f(\bm{x})-\mathbb{E}_{P_{y_c}}f(\bm{x})\right|
	\leq L\sum_i W_i(y_c)|y_i-y_c|.
\end{equation}
Taking the supremum over $f\in\mathcal{F}$ proves Eq.~\eqref{eq:vicinal_bias_bound}. \hfill$\square$

\noindent\textbf{Proof of Proposition~\ref{prop:neff_variance}.}
Let $a_i(y_c)\geq0$, $A(y_c)=\sum_i a_i(y_c)>0$, and
$\bar W_i(y_c)=a_i(y_c)/A(y_c)$. Then
$\sum_i\bar W_i(y_c)=1$ and
$N_{\mathrm{eff}}(y_c;a)=1/\sum_i\bar W_i^2(y_c)$ by
Eq.~\eqref{eq:neff_def}. Condition on the target label $y_c$, the
normalized weights, and the current network parameters. Under the
independence assumption for the scalar terms $\ell_i(y_c)$,
\begin{equation}
	\mathrm{Var}\!\left[\widehat L_D(y_c)\right]
	=
	\mathrm{Var}\!\left[\sum_i \bar W_i(y_c)\ell_i(y_c)\right]
	=
	\sum_i \bar W_i^2(y_c)\mathrm{Var}[\ell_i(y_c)].
\end{equation}
Therefore,
\begin{equation}
	\begin{aligned}
		\mathrm{Var}\!\left[\widehat L_D(y_c)\right]
		&\leq
		\max_i \mathrm{Var}[\ell_i(y_c)]
		\sum_i \bar W_i^2(y_c)\\
		&=
		\frac{\max_i \mathrm{Var}[\ell_i(y_c)]}
		{N_{\mathrm{eff}}(y_c;a)}.
	\end{aligned}
\end{equation}
This proves Eq.~\eqref{eq:neff_variance}. The result is conditional on
the fixed target label, fixed weights, and fixed network state. It is
therefore an estimator-level variance calculation, rather than a claim
about the full stochastic dynamics of GAN training. \hfill$\square$

\noindent\textbf{Proof of Proposition~\ref{prop:hav_condition_mass}.}
Let $d_i=|y_i-y_c|$ and let $S=\{i:d_i\leq\kappa_{y_c}\}$ be the retained adaptive-vicinity set. Since HAV uses the gated weights without renormalization, its retained mass is
\begin{equation}
	Z_{\mathrm{HAV}}=\sum_i\widetilde{W}_{i,y_c}
	=\sum_{i\in S}W_i.
\end{equation}
Because $W_i\geq0$ and $\sum_i W_i=1$, immediately $0\leq Z_{\mathrm{HAV}}\leq1$.

The raw HAV weighted condition-mismatch mass is
\begin{equation}
	\mathcal{C}_{\mathrm{HAV}}^{\mathrm{raw}}
	=\sum_i\widetilde{W}_{i,y_c}d_i
	=\sum_{i\in S}W_i d_i,
\end{equation}
whereas the corresponding SAV weighted condition-mismatch mass is
\begin{equation}
	\mathcal{C}_{\mathrm{SAV}}
	=\sum_i W_i d_i
	=\sum_{i\in S}W_i d_i+\sum_{i\notin S}W_i d_i.
\end{equation}
Therefore,
\begin{equation}
	\mathcal{C}_{\mathrm{SAV}}
	-\mathcal{C}_{\mathrm{HAV}}^{\mathrm{raw}}
	=\sum_{i\notin S}W_i d_i
	\geq0,
\end{equation}
which proves Eq.~\eqref{eq:hav_condition_mismatch_bound}. \hfill$\square$

\noindent\textbf{Proof of Proposition~\ref{prop:dre_approx_error}.}
Let $e=\hat r-r$. The assumption $\|\hat r-r\|_{L_2(p_g)}\leq\varepsilon_\text{dre}$ means that, for the fixed condition $y$ considered in the proposition,
\begin{equation}
	\|\hat r-r\|_{L_2(p_g)}
	=
	\left(
	\mathbb{E}_{\bm{x}\sim p_g(\bm{x}|y)}
	\left[
	\left(\hat r(\bm{x}|y)-r(\bm{x}|y)\right)^2
	\right]
	\right)^{1/2}
	\leq \varepsilon_\text{dre},
\end{equation}
or equivalently $\|e\|_{L_2(p_g)}\leq\varepsilon_\text{dre}$. The target Pearson divergence can be written as
\begin{equation}
	\mathrm{Div}_{\chi^2}(p_r\|p_g)=\mathbb{E}_{p_g}[(r-1)^2].
\end{equation}
We then expand the difference between the estimated and true generator penalties:
\begin{align}
	\left|\mathbb{E}_{p_g}[(\hat r-1)^2-(r-1)^2]\right|
	&=\left|\mathbb{E}_{p_g}[e(\hat r+r-2)]\right| \nonumber\\
	&=\left|\mathbb{E}_{p_g}[e(2(r-1)+e)]\right|.
\end{align}
Applying the triangle inequality and Cauchy-Schwarz gives
\begin{align}
	\left|\mathbb{E}_{p_g}[e(2(r-1)+e)]\right|
	&\leq 2\left|\mathbb{E}_{p_g}[e(r-1)]\right|+\mathbb{E}_{p_g}[e^2] \nonumber\\
	&\leq 2\|e\|_{L_2(p_g)}\|r-1\|_{L_2(p_g)}+\|e\|_{L_2(p_g)}^2.
\end{align}
Finally, using $\|e\|_{L_2(p_g)}\leq\varepsilon_\text{dre}$ and $\|r-1\|_{L_2(p_g)}=\sqrt{\mathrm{Div}_{\chi^2}(p_r\|p_g)}$ gives Eq.~\eqref{eq:dre_approx_error}. \hfill$\square$

\end{document}